\documentclass{article}

\PassOptionsToPackage{numbers,compress}{natbib}
\usepackage[preprint]{neurips_2026}

\usepackage[utf8]{inputenc}
\usepackage[T1]{fontenc}
\usepackage{hyperref}
\usepackage{url}
\usepackage{booktabs}
\usepackage{amsfonts}
\usepackage{amsmath}
\usepackage{amssymb}
\usepackage{nicefrac}
\usepackage{microtype}
\usepackage{xcolor}
\usepackage{graphicx}
\usepackage{multirow}
\usepackage{enumitem}
\usepackage{algorithm}
\usepackage{algorithmic}
\usepackage{subcaption}
\usepackage{wrapfig}
\usepackage{bm}

\title{Symbolic Regression via Latent Iterative Refinement}

\author{
  Xieting Chu \\
  Georgia Institute of Technology \\
  \And
  Sriram Vishwanath \\
  Georgia Institute of Technology \\
  \And
  Vijay Ganesh \\
  Georgia Institute of Technology
}

\begin{document}

\maketitle

\begin{abstract}
Symbolic regression (SR) seeks closed-form mathematical expressions that fit observed data.
Neural SR methods amortize the search by training an encoder to map observations directly to expressions in a single pass, but this amortized inference leaves a residual \emph{amortization gap} between its one-shot prediction and the true posterior.
We propose \textbf{Latent Equation Embedding (LEE)}, a framework that closes this gap through \emph{iterative amortized inference} in a functionally-grounded latent space.
LEE learns a shared latent space $\mathcal{Z}$ equipped with three components:
(1)~an encoder $f_\theta$ that jointly embeds symbolic tokens and numerical observations into a single latent vector $\bm{z}$;
(2)~an expression decoder $g_\text{expr}$ that reconstructs formulas from $\bm{z}$;
and (3)~an evaluation decoder $g_\text{eval}$ that predicts function values from $\bm{z}$, explicitly grounding the latent space in functional behavior.
At inference, LEE performs iterative refinement:
$\bm{z}_{t+1} = f_\theta\bigl(g_\text{expr}(\bm{z}_t),\;\mathcal{D}\bigr)$,
re-encoding decoded expressions jointly with observations to progressively improve the latent estimate.
LEE uses the encoder itself as a learned inference optimizer: each re-encoding step implicitly computes the mismatch between the candidate and the data. Because $g_\text{eval}$ is differentiable in $\bm{z}$, we additionally interleave continuous gradient descent with the discrete re-encoding, yielding a hybrid \emph{iterative\,+\,gradient refinement}.

On SRBench across three noise levels, against $19$ baselines spanning GP, symbolic--neural hybrids, and pre-trained Transformers, LEE produces expressions $2$--$10{\times}$ simpler than the strongest accuracy-oriented baselines---Operon, GP-GOMEA, TPSR, RAG-SR, and GenSR (complexity $8$--$11$ vs.\ $20$--$90$)---advancing the low-complexity region of the accuracy--complexity Pareto frontier and degrading gracefully as noise grows.
\end{abstract}

\section{Introduction}
\label{sec:intro}

Symbolic regression (SR) recovers interpretable mathematical expressions from data.
Formally, given observations $\mathcal{D} = \{(\bm{x}_i, y_i)\}_{i=1}^{N}$, the goal is to find $e^* = \arg\max_{e \in \mathcal{E}} p(e \mid \mathcal{D})$, where $\mathcal{E}$ is the space of symbolic expressions---a mixed search space whose skeleton (operators, variables, tree structure) is discrete and combinatorial while the embedded real-valued constants are continuous~\citep{schmidt2009distilling, udrescu2020ai, cranmer2023interpretable}.

\noindent\textbf{The amortization gap and existing remedies.}
Neural SR methods~\citep{biggio2021neural, kamienny2022end} replace the hours-long combinatorial search of classical genetic programming~\citep{schmidt2009distilling, cranmer2023interpretable} with an amortized inference model $f_\phi$ that maps observations directly to approximate posterior parameters $\bm{\lambda}^{(i)} \leftarrow f_\phi(\bm{x}^{(i)})$ (e.g.\ the logits of an autoregressive decoder). This one-shot prediction is fast but inherently limited: \citet{marino2018iterative} shows that the gap between the amortized estimate and the true optimum---the \emph{amortization gap}---grows with problem complexity; E2ESR~\citep{kamienny2022end} drops from $R^2{=}0.857$ on Feynman to $R^2{=}0.361$ on out-of-distribution black-box data. A second line of work introduces a latent space that pairs numerical and symbolic representations: SNIP~\citep{meidani2023snip} learns a discriminative pairing via contrastive pre-training (which cannot be searched directly), while GenSR~\citep{li2026gensr} learns a generative latent space via a dual-branch CVAE and refines the prior estimate with CMA-ES~\citep{hansen2001completely}. GenSR's CMA-ES is effective, but it treats the learned model as a black-box optimization objective: it uses only scalar fitness evaluations, discards the encoder's structural information, scales as $O(d_z^2)$, and carries no semantic understanding of why a candidate fits poorly. We close the amortization gap with a different lever: a learned latent search that exploits the encoder's own forward pass.

\noindent\textbf{Our approach: iterative amortized inference.}
\citet{marino2018iterative} show that the amortization gap can be closed by \emph{iterative inference models} that learn to optimize approximate posteriors by encoding gradients:
\begin{equation}
\bm{\lambda}_{t+1}^{(i)} \leftarrow f_t\bigl(\nabla_{\bm{\lambda}} \mathcal{L}_t^{(i)},\; \bm{\lambda}_t^{(i)};\; \phi\bigr).
\label{eq:iterative-amortized}
\end{equation}
LEE instantiates this principle in the symbolic regression setting.
Rather than encoding explicit gradients, LEE uses the decoded expression itself as an implicit error signal (an instance of the error-encoding variant of iterative amortized inference).
The encoder, when given both a candidate expression $\hat{e}_t = g_\text{expr}(\bm{z}_t)$ and the observations $\mathcal{D}$, can internally attend to the mismatch between the expression's predictions and the data---effectively computing a \emph{learned residual}.
This yields the LEE update rule:
\begin{equation}
\boxed{\bm{z}_{t+1} = f_\theta\bigl(\hat{e}_t,\; \mathcal{D}\bigr), \qquad \hat{e}_t = g_\text{expr}(\bm{z}_t)}
\label{eq:lee-update}
\end{equation}
Each iteration refines $\bm{z}$ by re-encoding a decoded expression jointly with observations, closing the amortization gap through the model's own inference pathway.

\noindent\textbf{Contributions.} We introduce three design choices that together enable iterative amortized inference for SR:
\begin{enumerate}[leftmargin=1.5em,itemsep=2pt]
\item \textbf{Iterative latent search via encode--decode--reencode} (Section~\ref{sec:iterative}).
The update rule (\ref{eq:lee-update}) uses the model's own encoder as a learned inference optimizer, maintaining a candidate pool for diversity.
Unlike GenSR's CMA-ES, each step is semantically informed: the encoder attends to both the candidate's tokens and the observations.

\item \textbf{Evaluation decoder for functional grounding} (Section~\ref{sec:eval-decoder}).
A dedicated decoder $g_\text{eval}(\bm{z}, \bm{x}) \to \hat{y}$ forces the latent space to encode what a function does, not just its syntactic form.
This creates a latent geometry where proximity reflects functional similarity---a prerequisite for meaningful iterative search.

\item \textbf{Hybrid iterative $+$ gradient refinement} (Section~\ref{sec:iterative}c).
Because $g_\text{eval}$ is differentiable in $\bm{z}$, continuous gradient descent can interleave with the discrete re-encoding, yielding a search that is more robust to noise than either alone; its importance is verified by a dedicated ablation (Section~\ref{sec:ablation}).
\end{enumerate}

On the SRBench benchmark suite~\citep{la2021contemporary}, LEE produces expressions $2$--$10{\times}$ simpler than the strongest accuracy-oriented baselines (Operon, GP-GOMEA, TPSR, RAG-SR, GenSR) while advancing the low-complexity region of the accuracy--complexity Pareto frontier.

\section{Background and Related Work}
\label{sec:related}

\noindent\textbf{Direct search in $\mathcal{E}$.}
Genetic programming (GP) methods sample and mutate expression trees directly in the discrete skeleton space, delegating constant fitting to an inner numeric optimizer. Contemporary high-performance GP systems such as Operon~\citep{burlacu2020operon}, GP-GOMEA~\citep{virgolin2021improving}, PySR~\citep{cranmer2023interpretable}, and Bingo~\citep{randall2022bingo} drive much of the state-of-the-art on SRBench. GP achieves high accuracy but requires repeated per-dataset evolutionary search, incurring wall-clock costs that scale with both dataset and population sizes.

\noindent\textbf{Amortized inference (one-shot).}
Pre-trained neural SR models amortize the per-dataset search with a single forward pass of an autoregressive decoder: NeSymReS~\citep{biggio2021neural} and E2E-SR~\citep{kamienny2022end} train a direct mapping $f_\phi: \mathcal{D} \to \hat{e}$. A second family combines amortized components with search-style augmentations, including DSR~\citep{petersen2019deep}, TPSR~\citep{shojaee2023transformer}, uDSR~\citep{landajuela2022unified}, and RAG-SR~\citep{zhang2025rag}. These models are fast at inference but incur a residual amortization gap: the one-shot estimate rarely matches the true posterior, and the gap widens on distributions the encoder did not see during pre-training.

\noindent\textbf{Latent-space methods.}
Another line of work learns a latent space that pairs numerical and symbolic representations and uses it either as an initialization or as the explicit search domain. SNIP~\citep{meidani2023snip} aligns numerical data and symbolic expressions via contrastive pre-training; the resulting discriminative embedding captures shared structure but is not generative, so it typically seeds a downstream decoder rather than acting as a search domain itself. GenSR~\citep{li2026gensr} instead learns a generative latent space via a dual-branch CVAE, framing SR as maximizing $p(\text{Equ.} \mid \text{Num.})$ through the ELBO
\begin{equation}
\log p(F \mid X) \geq \mathbb{E}_{q(z \mid X,F)}\bigl[\log p(F \mid X, z)\bigr] - D_\text{KL}\bigl(q(z \mid X,F) \;\|\; p(z \mid X)\bigr),
\label{eq:gensr-elbo}
\end{equation}
where the posterior branch encodes both expression $F$ and numerical data $X$ into $q(z \mid X, F)$ and the prior branch encodes only $X$ into $p(z \mid X)$. At inference, CMA-ES refines the prior-branch estimate in $\mathcal{Z}$ using only scalar fitness feedback, so the search is gradient-free w.r.t.\ the model and discards the encoder--decoder's structural information.

\noindent\textbf{Iterative amortized inference.}
\citet{marino2018iterative} propose closing the amortization gap by learning to iteratively refine approximate posteriors (Eq.~\ref{eq:iterative-amortized}); standard amortized inference is the one-step $t{=}0$ special case. The principle has not been applied to symbolic regression because (i) the latent space must be functionally meaningful for a refinement step to translate into a better expression, and (ii) the encoder must be trained to consume its own decoded outputs as input---two conditions that the LEE design explicitly satisfies.

\section{Method: Latent Equation Embedding}
\label{sec:method}

\begin{figure}[t]
\centering
\includegraphics[width=\linewidth]{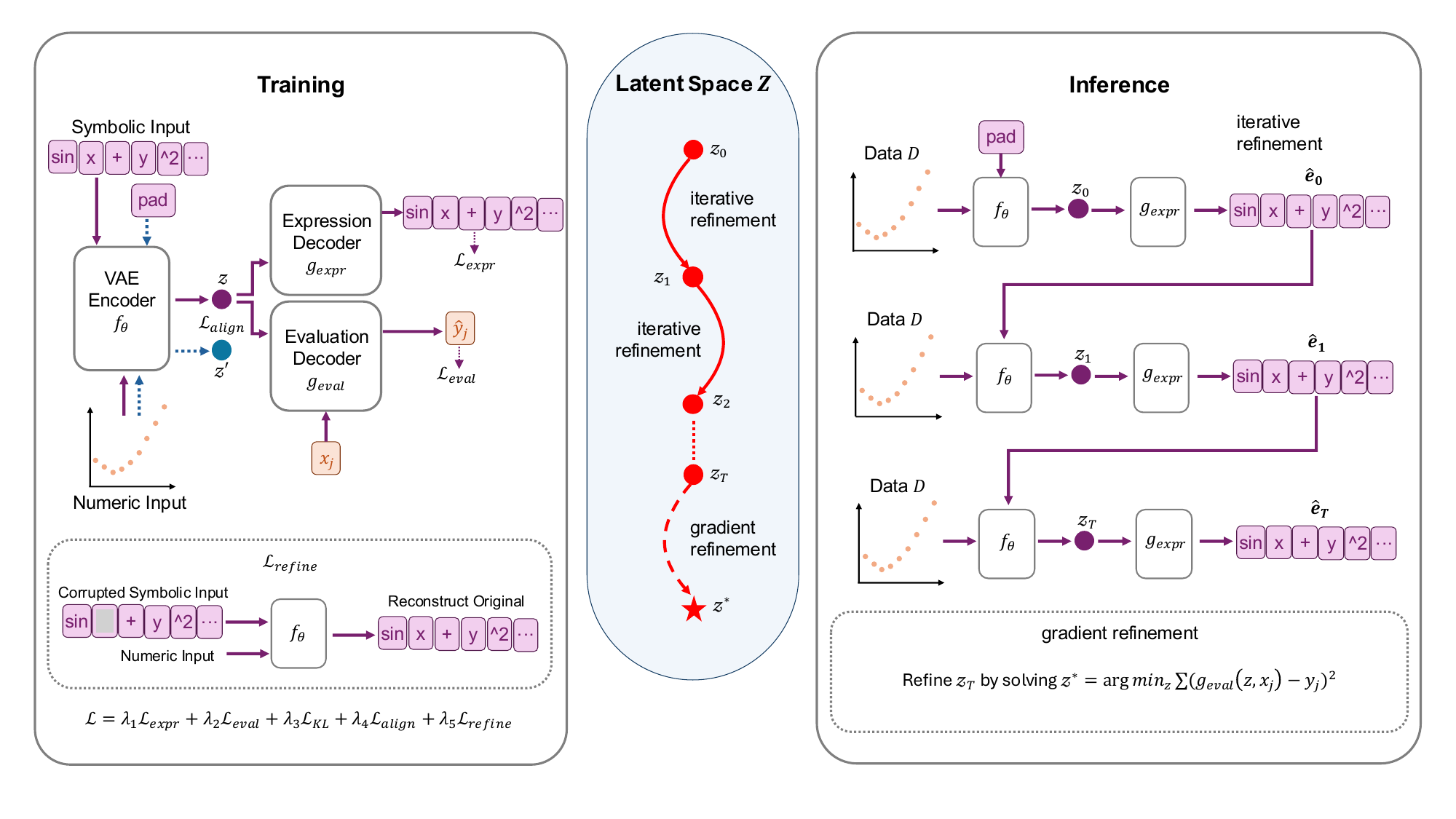}
\caption{\textbf{LEE architecture.}
\textbf{Left (training).} The encoder $f_\theta$ maps an expression and its scatter $\mathcal D$ to a latent $\bm z$, from which $g_\text{expr}$ reconstructs tokens and $g_\text{eval}$ predicts function values at queries $\bm x_j$; the five training losses (Sec.~\ref{sec:training}) are labeled at the points where they apply. The dashed path repeats the encode with tokens replaced by \texttt{pad} to define the scatter-only branch used by $\mathcal L_\text{align}$; the boxed inset depicts $\mathcal L_\text{refine}$ (random token corruption $\to$ denoised reconstruction).
\textbf{Middle.} Inference trajectory in $\mathcal Z$: solid red steps are discrete iterative refinement, the dashed segment is the final gradient refinement to $\bm z^{*}$.
\textbf{Right (inference).} Starting from $\bm z_0=f_\theta(\texttt{pad},\mathcal{D})$, each step decodes $\hat{\bm e}_t=g_\text{expr}(\bm z_t)$ and re-encodes $(\hat{\bm e}_t,\mathcal{D})$ to $\bm z_{t+1}$ (Eq.~\ref{eq:lee-update}); after $T$ steps, gradient refinement on $g_\text{eval}$ produces $\bm z^{*}$.}
\label{fig:framework}
\end{figure}

LEE consists of three jointly trained components sharing a latent space $\mathcal{Z} \subset \mathbb{R}^{d_z}$ (Figure~\ref{fig:framework}):
\begin{align}
\text{Encoder:} \quad & \bm{z} = f_\theta(\bm{t}, \mathcal{D}) \in \mathbb{R}^{d_z}, \label{eq:encoder} \\
\text{Expression decoder:} \quad & \hat{\bm{t}} = g_\text{expr}(\bm{z}), \label{eq:expr-dec} \\
\text{Evaluation decoder:} \quad & \hat{y}(\bm{x}) = g_\text{eval}(\bm{z}, \bm{x}), \label{eq:eval-dec}
\end{align}
where $\bm{t} \in \mathcal{V}^L$ is the symbolic token sequence and $\mathcal{D}$ is a set of scatter observations.
At inference, $\bm{t}$ is unavailable; the initial encoding uses only $\mathcal{D}$.

\subsection{Encoder}
\label{sec:encoder}

The encoder takes two input streams:

\noindent\textbf{Symbolic stream.} Each token $t_j$ of the expression is embedded as $\bm{h}_j^\text{sym} = \text{Embed}(t_j) \in \mathbb{R}^d$.

\noindent\textbf{Numeric stream.} Each observation $(\bm{x}_i, y_i)$ is embedded by a two-layer MLP, producing $\bm{h}_i^\text{num} \in \mathbb{R}^d$. Coordinates and function values are both log-compressed by $\tilde{u} = \text{sign}(u)\cdot\log(1+|u|)$ for numerical stability across many orders of magnitude. Non-finite values are handled by learnable special embeddings.

\noindent\textbf{Fusion.} Symbolic and numeric embeddings are concatenated along the sequence dimension, processed by an $L$-layer Transformer~\citep{vaswani2017attention}, and masked mean-pooled into a single vector $\bm{h}=\text{MeanPool}(\text{Transformer}([\bm{h}^\text{sym};\bm{h}^\text{num}]))$. Two linear heads map $\bm h$ to Gaussian parameters $\bm\mu=W_\mu\bm h$ and $\log\bm\sigma^2=W_\sigma\bm h$, and the latent vector is drawn by reparameterization~\citep{kingma2013auto}:
\begin{equation}
\bm{z} = \bm{\mu} + \bm{\sigma}\odot\bm{\epsilon},\qquad \bm{\epsilon}\sim\mathcal{N}(\bm{0},\bm{I}).
\label{eq:encoder-pool}
\end{equation}
The Gaussian posterior $q(\bm z\mid\bm t,\mathcal D)=\mathcal N(\bm\mu,\,\text{diag}(\bm\sigma^2))$ is regularized toward the standard-normal prior by $\mathcal{L}_\text{KL}$ (Sec.~\ref{sec:training}).

\subsection{Expression Decoder}

The expression decoder autoregressively generates prefix-notation tokens from $\bm{z}$.
The latent vector is projected into $K$ \emph{memory tokens} that serve as cross-attention keys:
\begin{equation}
\bm{M}_\text{expr} = \text{reshape}(W_m \bm{z}) \in \mathbb{R}^{K \times d_\text{expr}}.
\label{eq:memory}
\end{equation}
A causal Transformer decoder cross-attends to $\bm{M}_\text{expr}$, producing logits at each step:
$p_\theta(t_j \mid \bm{z}, t_{<j}) = \text{softmax}\bigl(W_o \cdot \text{TransDec}(\bm{M}_\text{expr}, t_{<j})\bigr)$.

\subsection{Evaluation Decoder: Grounding the Latent Space}
\label{sec:eval-decoder}

The evaluation decoder predicts function values at arbitrary query coordinates $\{\bm{q}_j\}_{j=1}^M$ from $\bm{z}$:
\begin{equation}
\bm{M}_\text{eval} = \text{reshape}(W_e \bm{z}) \in \mathbb{R}^{K \times d_\text{eval}}, \qquad \hat{y}_j = \text{MLP}\bigl(\text{TransDec}(\bm{M}_\text{eval}, \text{MLP}(\bm{q}_j))\bigr).
\label{eq:eval-decoder}
\end{equation}

\noindent\textbf{Why an evaluation decoder?}
Consider two expressions $e_1 = 2\sin(x)\cos(y)$ and $e_2 = \sin(x+y) + \sin(x-y)$. They are syntactically distant --- one is a scalar multiple of a product of two trig atoms, the other is a sum of trig functions applied to compound arguments --- but functionally identical by the product-to-sum identity.
Without $g_\text{eval}$, no loss explicitly ties $\bm{z}$ to function values: symbolic reconstruction penalizes $e_1$ and $e_2$ equally for being decoded as each other, and while the scatter input together with $\mathcal{L}_\text{align}$ provides an indirect pressure toward numerical consistency, nothing forces functionally-equivalent expressions to map to nearby $\bm{z}$.
With $g_\text{eval}$, the latent space must encode functional behavior: for $e_1$ and $e_2$ to both decode to identical $\hat y$, they are pushed toward nearby $\bm{z}$. This encourages the soft equivalence
\begin{equation}
\bm{z}_1 \approx \bm{z}_2 \quad \text{whenever} \quad e_1(\bm{x}) = e_2(\bm{x}) \;\; \forall \bm{x},
\label{eq:functional-equiv}
\end{equation}
shaping a latent geometry in which proximity reflects functional similarity.
This is the geometry needed for iterative search: moving $\bm{z}$ toward a functionally better region tends to decode a functionally better expression. Because $g_\text{eval}$ is differentiable in $\bm{z}$, it also provides the gradient signal that enables the continuous refinement mode of Sec.~\ref{sec:iterative}(b).

\subsection{Training Objective}
\label{sec:training}

The total loss combines five terms:
\begin{equation}
\mathcal{L} = \lambda_\text{expr}\mathcal{L}_\text{expr} + \lambda_\text{eval}\mathcal{L}_\text{eval} + \lambda_\text{KL}\mathcal{L}_\text{KL} + \lambda_\text{align}\mathcal{L}_\text{align} + \lambda_\text{refine}\mathcal{L}_\text{refine}.
\label{eq:total-loss}
\end{equation}

\noindent\textbf{Expression reconstruction ($\mathcal{L}_\text{expr}$).} Cross-entropy over non-padding tokens:
$\mathcal{L}_\text{expr} = -\frac{1}{|\mathcal{T}|}\sum_{j \in \mathcal{T}} \log p_\theta(t_j \mid \bm{z}, t_{<j})$.

\noindent\textbf{Evaluation loss ($\mathcal{L}_\text{eval}$).} Scale-invariant MAE:
$\mathcal{L}_\text{eval} = \frac{1}{|\mathcal{V}_\text{fin}|} \sum_{j \in \mathcal{V}_\text{fin}} \frac{|\hat{y}_j - y_j|}{\max(|y_j|, 1)}$.

\noindent\textbf{Latent regularization ($\mathcal{L}_\text{KL}$).} Standard VAE KL divergence between the posterior $q(\bm{z} \mid \bm{t}, \mathcal{D})$ produced by the encoder's $(\bm\mu, \bm\sigma)$ head and a unit Gaussian prior $\mathcal{N}(\bm{0}, \bm{I})$, with a small weight $\lambda_\text{KL}$ to avoid posterior collapse.

\noindent\textbf{Cross-modal alignment ($\mathcal{L}_\text{align}$).}
At inference, only scatter is available. To bridge this modality gap we run the same encoder $f_\theta$ twice per training sample---once on the full (tokens, scatter) input, producing a posterior $q(\bm z \mid \bm t, \mathcal{D})$, and once on scatter alone (symbolic stream filled with \texttt{[pad]}), producing a scatter-only distribution $p(\bm z \mid \mathcal{D})$---and align the two via a conditional KL:
\begin{equation}
\mathcal{L}_\text{align} = D_\text{KL}\!\left(\,q(\bm z\mid \bm t, \mathcal{D})\,\big\|\,\text{sg}\!\left[p(\bm z\mid\mathcal{D})\right]\,\right).
\label{eq:align}
\end{equation}
Here $\text{sg}[\cdot]$ denotes the standard stop-gradient operator \citep{van2017neural,grill2020bootstrap}: during the backward pass it treats its argument as a constant, so the gradient of $\mathcal{L}_\text{align}$ flows only into the posterior ($q$) branch, while the scatter-only ($p$) branch is held fixed as the target. This matches the KL direction in GenSR's ELBO (Eq.~\ref{eq:gensr-elbo}) but is used as a standalone weighted term rather than part of a probabilistic objective.

\noindent\textbf{Iterative refinement ($\mathcal{L}_\text{refine}$).}
To train the encoder for the iterative regime, we simulate the inference-time loop during training.
Expression tokens are randomly corrupted (drops, swaps, substitutions), producing a noisy expression $\tilde{e}$.
The encoder must map $(\tilde{e}, \mathcal{D})$ to a $\bm{z}$ that decodes to the \emph{original} expression:
\begin{equation}
\mathcal{L}_\text{refine} = -\frac{1}{|\mathcal{T}|}\sum_{j \in \mathcal{T}} \log p_\theta\bigl(t_j \mid f_\theta(\tilde{e}, \mathcal{D}),\; t_{<j}\bigr).
\label{eq:refine}
\end{equation}
This trains the encoder to act as a denoising inference optimizer~\citep{vincent2008extracting}: given a corrupted expression and the data, it must ``correct'' the latent representation.
At inference, decoded expressions play the role of $\tilde{e}$---they are imperfect approximations that the encoder refines.

\subsection{Iterative Latent Search}
\label{sec:iterative}

The inference procedure instantiates Eq.~\ref{eq:lee-update} through three complementary refinement strategies: \emph{iterative refinement} (discrete re-encoding through $f_\theta$, operating over a candidate pool), \emph{gradient refinement} (continuous descent in $\mathcal{Z}$ through the evaluation decoder), and their combination \emph{iterative\,+\,gradient refinement}. These form the core of our method and the basis of the ablation study in Sec.~\ref{sec:ablation}.

\noindent\textbf{Initialization.}
The initial latent vector encodes only the observations:
\begin{equation}
\bm{z}_0 = f_\theta(\varnothing, \mathcal{D}).
\label{eq:z0}
\end{equation}
From $\bm{z}_0$, we decode $n_\text{init}$ candidate expressions via greedy and temperature-sampled decoding, score each by $R^2 - \alpha \cdot \text{complexity}$, and keep the top $P$ as the initial pool $\Pi_0$.

\noindent\textbf{(a) Iterative refinement.}
At each step $t$, we sample a parent expression $\hat{e}_t^{(i)}$ from $\Pi_t$ with rank-weighted probability, and apply the update:
\begin{equation}
\bm{z}_{t+1}^{(i)} = f_\theta\bigl(\hat{e}_t^{(i)},\; \mathcal{D}\bigr), \qquad \hat{e}_{t+1}^{(j)} \sim g_\text{expr}\bigl(\bm{z}_{t+1}^{(i)}\bigr), \quad j = 1, \ldots, n_\text{new}.
\label{eq:iteration}
\end{equation}
New candidates are scored and merged into $\Pi_t$, keeping the top $P$ with complexity diversity. Constants in decoded expressions are refined via L-BFGS-B~\citep{byrd1995limited}. To prevent pool collapse, we periodically re-sample scatter points from $\mathcal{D}$ and decode fresh candidates from a new scatter-only $\bm{z}_0$ (every 5 batches; Sec.~\ref{sec:experiments}). This is the discrete, encoder-driven realization of Eq.~\ref{eq:lee-update}.

\noindent\textbf{(b) Gradient refinement.}
Because $g_\text{eval}(\bm{z}, \bm{x})$ is differentiable in $\bm{z}$, we can directly descend on the latent:
\begin{equation}
\bm{z}_{t+1} = \bm{z}_t - \eta\,\nabla_{\bm{z}}\bigl[\;\|\,g_\text{eval}(\bm{z}_t, \bm{X}) - \bm{y}\|_2^2 + \lambda_\text{prox}\,\|\bm{z}_t - \bm{z}_\text{anchor}\|_2^2\;\bigr].
\label{eq:grad-update}
\end{equation}
The proximal term keeps $\bm{z}_t$ near the decodable region of $\mathcal{Z}$. Every $d$ steps, we decode the current $\bm{z}_t$ and score the resulting expression; the best expression seen over the trajectory is returned. This mode exploits the functional grounding induced by $g_\text{eval}$ (Sec.~\ref{sec:eval-decoder}).

\noindent\textbf{(c) Iterative + gradient refinement.}
Iterative refinement excels at global exploration through discrete re-encoding; gradient refinement excels at local fine-tuning through continuous descent. The combined mode alternates between the two: every $d$ iterative refinement steps, we take the current pool champion $\bm{z}^*$, run $k$ steps of gradient descent (\ref{eq:grad-update}), decode, and merge the resulting expression back into the pool. A safety fallback uses the held-out validation fold: if the gradient step lowers validation $R^2$ relative to the pool champion, we revert to the pool champion. The test fold is untouched until final reporting. As we show in Sec.~\ref{sec:ablation}, this hybrid is especially valuable on noisy data, where gradient refinement locally denoises coefficients while iterative refinement maintains structural diversity through the pool.

\section{Experimental Setup}
\label{sec:experiments}

\noindent\textbf{Benchmarks.}
We evaluate on the SRBench benchmark suite~\citep{la2021contemporary}: \textbf{Strogatz} (14 ODE systems), \textbf{Feynman} (116 physics equations), and \textbf{black-box} (63 PMLB datasets without known ground truth). Ground-truth benchmarks are run at three target noise levels $\epsilon \in \{0, 0.01, 0.1\}$ (Gaussian noise with standard deviation proportional to the target range); black-box is noise-free.

\noindent\textbf{Data splits.}
We adopt SRBench's canonical 75\%/25\% train/test partition; the 25\% test fold matches SRBench's protocol exactly, making our test numbers directly comparable to published baselines. Internally, we carve a 20\% validation slice from the 75\% training portion, so each dataset is 60/15/25 train/val/test overall. $R^2$ on the test fold is the reported accuracy metric; the validation fold is used only for round selection within a trial and for the gradient-fallback decision, and is never observed by the model during search. Complexity is SymPy-simplified~\citep{meurer2017sympy} node count.

\noindent\textbf{Evaluation protocol.}
For each (dataset, $\epsilon$) pair, we run independent trials with distinct random seeds (data splits and search seeds) and report the mean $\pm$ standard deviation across trials. The main results in Table~\ref{tab:main} use $10$ trials per cell to match SRBench's published-baseline protocol; the ablation studies in Sec.~\ref{sec:ablation} and the appendix sensitivity sweeps use $3$ trials to keep the compute footprint manageable. Each trial is the best-of-$R{=}10$ rounds of iterative\,+\,gradient refinement (Sec.~\ref{sec:iterative}), with the winning round chosen by highest validation $R^2$. Aggregation at the dataset-group level uses the mean across datasets within each group.

\noindent\textbf{Training data.}
${\sim}13.4$M synthetic expressions from a stochastic context-free grammar (15 operators, 1--10 variables), paired with 200 scatter points from $\mathcal{U}(-10,10)^k$.

\noindent\textbf{Model.}
Encoder: $d{=}768$, 6 layers, 12 heads, $d_z{=}512$ (${\approx}75$M params).
Expression decoder: $d{=}512$, 8 layers, 8 heads, $K{=}4$ memory tokens (${\approx}50$M).
Evaluation decoder: $d{=}512$, 4 layers, 8 heads, $K{=}4$ (${\approx}25$M).
Total: ${\approx}150$M parameters. Training: AdamW~\citep{loshchilov2017decoupled} with cosine decay, single NVIDIA GH200.

\noindent\textbf{Inference.}
Pool size $P{=}16$, $n_\text{init}{=}32$, $T{=}200$ iterations per round, $n_\text{new}{=}3$ per iteration, batch $k{=}5$ parents processed together, scatter refresh every 5 batches, L-BFGS-B ramped $100\!\to\!300$ steps. Candidates are scored by $R^2_\text{train} - \alpha\cdot C(e)$ with $\alpha{=}0.002$ and $C(e)$ the SymPy-simplified node count; decoder sampling uses temperature $\tau{=}0.7$. The pool maintains \emph{complexity diversity} by keeping at most $\lceil P/4 \rceil$ candidates per complexity bucket, so that short and long expressions are both retained. For the combined mode, we insert one gradient segment ($k{=}50$ steps, $\eta{=}5\!\times\!10^{-3}$, $\lambda_\text{prox}{=}0.1$) every 25 iterative refinement steps. Rounds run 8-way parallel on a single GH200 node.

\noindent\textbf{Baselines.}
We compare against 19 SRBench methods: GP-based (Operon, GP-GOMEA, SBP-GP, GPlearn, AFP, AFP-FE, EPLEX, ITEA), symbolic + deep hybrids (DSR, RSRM, MDL, SPL, AIFeynman2), and neural SR (NeurSR, E2ESR, SNIP, TPSR, RAG-SR, GenSR). Baseline numbers are taken from the published GenSR paper~\citep{li2026gensr} and the SRBench 2.0 feather data where applicable, which follow an identical 10-trial, 75/25 split, $R^2$ test-fold protocol. Since LEE completes per-dataset in tens of seconds---well below the SRBench compute budget that bounds the baselines (Appendix~\ref{app:timing})---our $R^2$ and complexity are hardware-agnostic and directly comparable to the published values.

\section{Results}
\label{sec:results}

Our empirical study answers three questions: (i) How does LEE compare to existing SR methods across noise levels and benchmark types, both in headline metrics and on the accuracy--complexity Pareto frontier (Sec.~\ref{sec:overall}--\ref{sec:pareto})? (ii) Which refinement strategies (iterative, gradient, or their combination) are responsible for the results (Sec.~\ref{sec:ablation})? (iii) Does the iterative re-encoding update specifically---rather than the backbone, training data, or scoring---drive the accuracy gain over one-shot decoding and CMA-ES on the same checkpoint (Sec.~\ref{sec:same-backbone})? Iterative-convergence behavior in $\mathcal{Z}$ and detailed timing data are deferred to Appendix~\ref{sec:appendix-convergence} and Appendix~\ref{app:timing}.

\subsection{Overall Comparison}
\label{sec:overall}

Table~\ref{tab:main} reports mean test $R^2$ and simplified complexity across Strogatz, Feynman, and black-box for three noise levels. Both LEE and baseline numbers follow SRBench's $10$-trial $75\%/25\%$ protocol on the same $25\%$ test fold; each LEE trial is a best-of-$10$-rounds run. LEE consistently occupies the low-complexity corner of the accuracy--complexity trade-off, and its accuracy degrades gracefully with noise on the ground-truth benchmarks, in contrast to several neural methods that sharply collapse.

\begin{table}[t]
\caption{\textbf{SRBench results across noise levels.} Mean test $R^2$ ($\uparrow$) and mean simplified complexity (Cmplx $\downarrow$). Black-box has no ground truth and is run noise-free. LEE numbers are means over $10$ trials, matching SRBench's published-baseline protocol; per-cell standard deviations are reported in Appendix~\ref{sec:appendix-std} for compactness. \textbf{Bold}: best neural/hybrid; \underline{underline}: best overall.}
\label{tab:main}
\centering
\small
\setlength{\tabcolsep}{3pt}
\begin{tabular}{l cc cc cc cc cc cc cc}
\toprule
& \multicolumn{6}{c}{\textbf{Strogatz (14)}} & \multicolumn{6}{c}{\textbf{Feynman (116)}} & \multicolumn{2}{c}{\textbf{Black-box (63)}} \\
\cmidrule(lr){2-7}\cmidrule(lr){8-13}\cmidrule(lr){14-15}
& \multicolumn{2}{c}{$\epsilon{=}0$} & \multicolumn{2}{c}{$\epsilon{=}0.01$} & \multicolumn{2}{c}{$\epsilon{=}0.1$}
& \multicolumn{2}{c}{$\epsilon{=}0$} & \multicolumn{2}{c}{$\epsilon{=}0.01$} & \multicolumn{2}{c}{$\epsilon{=}0.1$}
& \multicolumn{2}{c}{---} \\
Method & $R^2$ & C & $R^2$ & C & $R^2$ & C & $R^2$ & C & $R^2$ & C & $R^2$ & C & $R^2$ & C \\
\midrule
\multicolumn{15}{l}{\textit{Genetic programming}} \\
Operon     & .988 & 59  & .983 & 82  & .938 & 83  & .989 & 70  & .988 & 88  & .985 & 89  & .794 & 66 \\
GP-GOMEA   & .992 & 36  & .978 & 43  & .967 & 44  & \underline{.996} & 35  & \underline{.997} & 45  & \underline{.996} & 46  & .738 & 30 \\
SBP-GP     & .981 & 712 & .981 & 851 & .932 & 901 & .994 & 489 & .995 & 596 & .990 & 622 & .787 & 634 \\
GPlearn    & .769 & 29  & .796 & 31  & .823 & 26  & .881 & 72  & .889 & 60  & .891 & 49  & .539 & 19 \\
AFP        & .925 & 38  & .915 & 39  & .911 & 44  & .959 & 37  & .961 & 41  & .958 & 41  & .633 & 35 \\
AFP-FE     & .944 & 46  & .958 & 49  & .950 & 51  & .981 & 40  & .982 & 47  & .983 & 49  & .640 & 36 \\
EPLEX      & .812 & 50  & .856 & 53  & .882 & 54  & .987 & 53  & .991 & 54  & .990 & 46  & .737 & 53 \\
ITEA       & .792 & 11  & ---  & --- & .910 & 15  & ---  & --- & ---  & --- & ---  & --- & .629 & 117 \\
\midrule
\multicolumn{15}{l}{\textit{Symbolic--neural hybrid}} \\
DSR        & .760 & 16  & .820 & 18  & .809 & 18  & .844 & 15  & .878 & 16  & .878 & 16  & .562 & 10 \\
RSRM       & .550 & 13  & .597 & 14  & .555 & 14  & .800 & 13  & .809 & 13  & .810 & 13  & .332 & 9 \\
MDL        & .990 & 14  & .972 & 20  & .969 & 20  & .917 & 23  & .914 & 31  & .910 & 31  & .626 & 30 \\
SPL        & .739 & 15  & .739 & 15  & .772 & 14  & .707 & 13  & .713 & 13  & .711 & 14  & .547 & 13 \\
AIFeynman2 & .646 & 22  & .775 & 32  & .317 & 24  & .931 & 124 & .873 & 155 & .225 & 177 & .211 & 2240 \\
\midrule
\multicolumn{15}{l}{\textit{Pre-trained neural}} \\
NeurSR     & .521 & 11  & .518 & 12  & .505 & 13  & .396 & 13  & .394 & 13  & .382 & 14  & .123 & 13 \\
E2ESR      & .534 & 32  & .503 & 36  & .515 & 38  & .857 & 36  & .834 & 40  & .771 & 44  & .361 & 61 \\
SNIP       & \textbf{\underline{.995}} & 29 & \textbf{.984} & 29 & .919 & 39  & .985 & 32  & .987 & 33  & .992 & 38  & .334 & 39 \\
TPSR       & .965 & 56  & .980 & 56  & .971 & 56  & .992 & 57  & .991 & 64  & .984 & 67  & ---  & --- \\
RAG-SR     & .991 & 46  & .987 & 49  & .969 & 46  & .993 & 46  & .990 & 72  & .985 & 75  & ---  & --- \\
GenSR      & \underline{.992} & 20  & \textbf{\underline{.994}} & 20  & \textbf{\underline{.977}} & 20  & .987 & 23  & .987 & 23  & .989 & 24  & \textbf{\underline{.842}} & 35 \\
\midrule
\textbf{LEE (ours)} & .854 & \textbf{8.1} & .876 & \textbf{8.9} & .880 & \textbf{8.3} & .884 & \textbf{9.9} & .884 & \textbf{10.1} & .824 & \textbf{10.6} & .559 & \textbf{9.0} \\
\bottomrule
\end{tabular}
\vspace{-4pt}
\end{table}

\noindent\textbf{Accuracy.} On Strogatz and Feynman, LEE lies within 6--14 $R^2$ points of the top GP methods, which is the cost of the simplicity trade-off discussed below; LEE is not an accuracy-SOTA method. Under noise, however, LEE's Strogatz $R^2$ rises slightly from $0.854$ to $0.880$ as $\epsilon$ grows from $0$ to $0.1$, whereas SNIP drops by $0.076$ and E2ESR by $0.019$ (while producing $\sim 4\times$ larger expressions); we hypothesize that input noise widens the encoder's posterior, increasing search diversity at no accuracy cost. On black-box (no ground truth, out-of-distribution), LEE reaches $R^2{=}0.559$, ahead of one-shot neural methods (SNIP $0.334$, E2ESR $0.361$) while keeping complexity $4$--$7\times$ smaller.

\noindent\textbf{Complexity.} Across all settings, LEE produces the simplest expressions---complexity $8$--$11$ versus $15$--$70{+}$ for all competing neural methods and most GP baselines---while retaining competitive $R^2$. This is the Pareto-differentiating property of LEE: where other methods trade accuracy against complexity with larger expressions, LEE advances the low-complexity region of the frontier.

\noindent\textbf{Speed.} LEE's inference is fast: one dataset completes in tens of seconds of wallclock time on a single GH200, faster than most GP baselines and within a small constant factor of the fastest one-shot neural methods. A detailed timing comparison is given in Appendix~\ref{app:timing}; because hardware varies substantially across baselines, we caution against over-interpreting absolute numbers.

\subsection{Pareto Analysis}
\label{sec:pareto}

\begin{figure}[t]
\centering
\begin{subfigure}[b]{0.48\textwidth}\includegraphics[width=\textwidth]{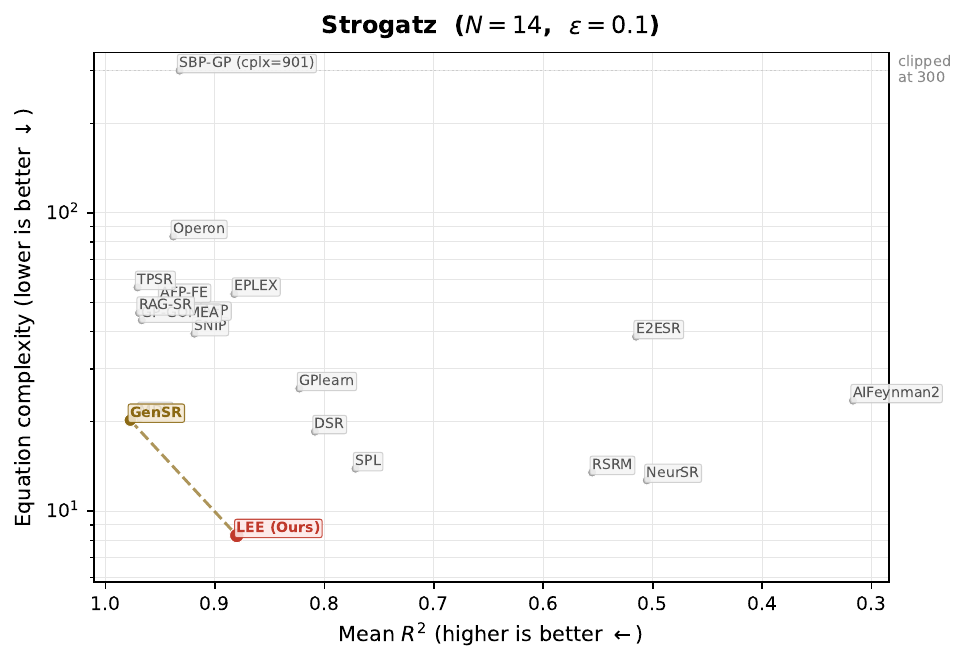}
\caption{Strogatz, $\epsilon{=}0.1$}\end{subfigure}\hfill
\begin{subfigure}[b]{0.48\textwidth}\includegraphics[width=\textwidth]{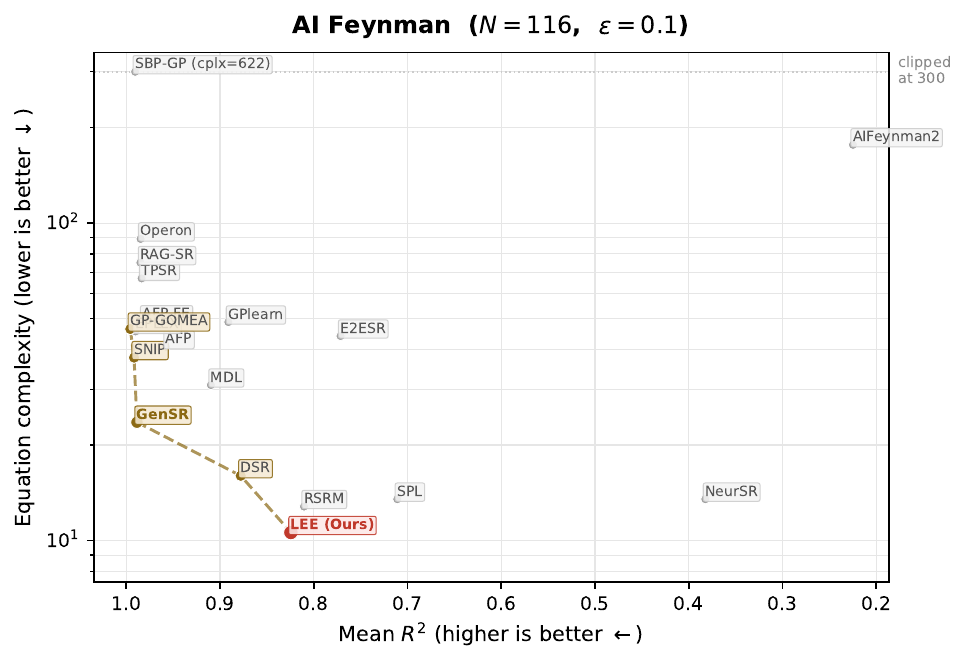}
\caption{Feynman, $\epsilon{=}0.1$}\end{subfigure}
\caption{\textbf{Pareto frontiers at $\epsilon{=}0.1$} (test $R^2$ vs.\ complexity, log-$x$). LEE sits in the low-complexity corner of the frontier, typically $2$--$7\times$ simpler than accuracy-comparable methods. The same qualitative picture holds at $\epsilon{=}0$ and $\epsilon{=}0.01$ (Appendix~\ref{sec:appendix-pareto}) and on black-box (Fig.~\ref{fig:pareto-bb}).}
\label{fig:pareto}
\end{figure}

Figure~\ref{fig:pareto} visualizes the same tables in the $(R^2,\,\text{complexity})$ plane. Three observations stand out:
(i) \textbf{Consistency across noise} (Appendix~\ref{sec:appendix-pareto}, Fig.~\ref{fig:pareto-clean}). LEE's position on the frontier is essentially invariant as $\epsilon$ grows from 0 to 0.1, while several neural methods (SNIP, TPSR, RAG-SR) move up and to the right (more complex, less accurate).
(ii) \textbf{Distinct regime.} No other method reaches complexity $<15$ at the accuracy LEE achieves; DSR and MDL are the closest competitors but sit at higher complexity or lower $R^2$.
(iii) \textbf{Black-box robustness} (Fig.~\ref{fig:pareto-bb}, Appendix~\ref{sec:appendix-pareto}). The black-box Pareto frontier is a staircase of non-dominated points at progressively higher complexity tiers---LEE and DSR both anchor the low-complexity end (cplx $\le 10$, $R^2 \approx 0.56$), then MDL ($30$), GP-GOMEA ($30$), and GenSR ($35$)---with LEE producing the simplest expressions on the frontier. One-shot neural baselines (SNIP, E2ESR, NeurSR) are dominated in both dimensions, reflecting how the single-pass inference distribution fails to transfer to OOD data.

\subsection{Why Combine Iterative and Gradient Refinement: Ablation Study}
\label{sec:ablation}

A central claim of this paper is that the combination of discrete encoder-driven iterative refinement and continuous gradient refinement via the evaluation decoder makes iterative amortized inference effective, particularly under noise. We test this with an ablation on Strogatz that disables each component:

\begin{table}[t]
\caption{\textbf{Search-strategy ablation on Strogatz} (mean $\pm$ std, each best-of-10-rounds with 200 iterations per round; $R=10$, $T=200$, 8-way parallel).}
\label{tab:ablation}
\centering
\small
\begin{tabular}{lcccc}
\toprule
& \multicolumn{2}{c}{$\epsilon{=}0$} & \multicolumn{2}{c}{$\epsilon{=}0.1$} \\
\cmidrule(lr){2-3}\cmidrule(lr){4-5}
Strategy & $R^2$ & Cmplx & $R^2$ & Cmplx \\
\midrule
iterative refinement ($f_\theta$ re-encoding)          & \textbf{0.872} $\pm$ 0.012 & 8.0 & 0.850 $\pm$ 0.024 & 9.5 \\
gradient refinement ($\nabla_{\bm z} g_\text{eval}$)   & 0.742 $\pm$ 0.027 & 9.2 & 0.744 $\pm$ 0.045 & 9.6 \\
\textbf{iterative + gradient refinement}               & 0.854 $\pm$ 0.005 & 8.1 & \textbf{0.880} $\pm$ 0.024 & \textbf{8.3} \\
\bottomrule
\end{tabular}
\vspace{-4pt}
\end{table}

\noindent\textbf{Gradient refinement alone is insufficient.}
Pure gradient refinement through $g_\text{eval}$ (row 2 of Table~\ref{tab:ablation}) lags iterative refinement by \textbf{13 $R^2$ points} at $\epsilon{=}0$ and \textbf{10 points} at $\epsilon{=}0.1$, and exhibits the highest run-to-run variance. The gradient signal moves $\bm{z}$ toward a \emph{training-loss} minimum, not toward a well-formed expression in $g_\text{expr}$'s decodable region, so without re-projection through the encoder, it overfits coefficients and drifts off-manifold.

\noindent\textbf{Iterative refinement alone is competitive at $\epsilon{=}0$ but loses accuracy under noise.}
Iterative refinement alone achieves the best $R^2$ on clean data (0.872), consistent with the idea that on noise-free problems, the discrete search space is well-structured and continuous refinement adds little. Under $\epsilon{=}0.1$, however, it drops to 0.850 with std $0.024$---a three-point accuracy hit and triple the variance of the combined mode at $\epsilon{=}0$.

\noindent\textbf{Combining both is the robust choice.}
Iterative\,+\,gradient refinement is within $0.02$ of iterative-only on clean data while having $2.4\times$ lower variance ($\pm 0.005$ vs.\ $\pm 0.012$), and it surpasses iterative-only on noisy data (0.880 vs.\ 0.850, a $+3\%$ gain), while also producing the simplest expressions (Cmplx 8.3 vs.\ 9.5). This matches our design intuition (Sec.~\ref{sec:iterative}c): iterative refinement maintains discrete diversity while gradient segments locally adjust $\bm{z}$ along the manifold shaped by $g_\text{eval}$ so that a subsequent decode lands on a better expression (whose constants L-BFGS-B then refines).

\subsection{Same-Backbone Search Comparison}
\label{sec:same-backbone}

To attribute LEE's gains specifically to the iterative re-encoding update---rather than to the backbone, training data, or scoring function---we compare three search procedures on the \emph{identical} pre-trained LEE checkpoint, identical scoring rule $s(e) = R^2 - \alpha C(e)$, and matched per-round decode budget of $n_\text{init} + T \cdot n_\text{new}$ decodes per round (Sec.~\ref{sec:appendix-inference}): \textbf{(a)} one-shot decode from $\bm z_0 = f_\theta(\texttt{pad}, \mathcal{D})$ (no search); \textbf{(b)} CMA-ES on $\bm z$ with $R^2$ fitness (GenSR-style; population $24$); \textbf{(c)} LEE iterative + gradient refinement (ours).

\begin{table}[t]
\caption{\textbf{Same-backbone search comparison} on Strogatz at $\epsilon{=}0.1$. All three procedures use the identical pre-trained LEE checkpoint, scoring, and matched decode budget. (a) and (b) are $3$-trial averages; (c) reports the headline $10$-trial number from Table~\ref{tab:main}.}
\label{tab:same-backbone}
\centering
\small
\begin{tabular}{lcc}
\toprule
Procedure & $R^2$ ($\uparrow$) & Cmplx ($\downarrow$) \\
\midrule
(a) One-shot decode from $\bm z_0$        & $0.795 \pm 0.019$ & $13.0 \pm 1.9$ \\
(b) CMA-ES on $\bm z$ ($R^2$ fitness)     & $0.848 \pm 0.011$ & $13.8 \pm 0.8$ \\
\textbf{(c) LEE iterative + gradient}     & $\mathbf{0.880 \pm 0.024}$ & $\mathbf{8.3 \pm 1.4}$ \\
\bottomrule
\end{tabular}
\vspace{-4pt}
\end{table}

Searching in $\bm z$ at all (a$\to$b) buys $+0.05$ $R^2$ over the one-shot baseline, confirming that the latent geometry is useful for search beyond the initial estimate. Replacing scalar-fitness CMA-ES with our encoder-driven iterative update (b$\to$c) buys another $+0.03$ $R^2$ and cuts complexity by ${\sim}40\%$. We note that CMA-ES is given $4{-}5\times$ the per-dataset wallclock budget of LEE iterative+gradient (its per-generation L-BFGS-B refinement is heavier than LEE's incremental pool update), and a longer budget would likely close part of the $R^2$ gap; the simplicity gap, however, is structural rather than budget-bound. The encoder's structured update therefore does two things that CMA-ES cannot match at any budget: (i) it produces simpler expressions, because the encoder's training distribution is biased toward simple skeletons; and (ii) it makes each step a single forward pass rather than a population evaluation with covariance updates and per-candidate constant refinement, so progress is incremental and unaffected by population synchronization.

\section{Conclusion and Future Work}
\label{sec:conclusion}

We presented LEE, a framework that casts symbolic regression as \emph{iterative amortized inference} in a functionally-grounded latent space.
The central equation, $\bm{z}_{t+1} = f_\theta(g_\text{expr}(\bm{z}_t), \mathcal{D})$, uses the model's own encoder as a learned inference optimizer, closing the amortization gap of one-shot methods like E2ESR and avoiding the black-box search of GenSR's CMA-ES.
A differentiable evaluation decoder further grounds $\mathcal{Z}$ in functional behavior, enabling both discrete re-encoding and continuous gradient-based refinement; our ablation (Sec.~\ref{sec:ablation}) shows that combining the two is essential under noise.
On SRBench across three noise levels, LEE occupies a distinctive Pareto position: $2$--$10\times$ simpler expressions than the strongest accuracy-oriented baselines (Operon, GP-GOMEA, TPSR, RAG-SR, GenSR) with accuracy within $0.10$--$0.17$ $R^2$ of those methods, modest wallclock cost, and graceful out-of-distribution behavior---well suited for scientific discovery where interpretability matters as much as fit.

\noindent\textbf{Future work.}
LEE's framework admits several natural extensions. Scaling the operator vocabulary and pre-training corpus should narrow the $0.10$--$0.17$ $R^2$ gap to top GP methods (Operon, GP-GOMEA) on clean benchmarks; coupling the same backbone with a higher-capacity generative prior (e.g., GenSR-style dual-branch encoding) for $\bm z_0$, and directly measuring latent distance between canonically-equivalent expressions to quantify the functional-grounding claim of Sec.~\ref{sec:eval-decoder}, would each tighten the framework further.

\begin{ack}
We thank our collaborators and colleagues for helpful discussions.
We thank our collaborators and colleagues for helpful discussions.
\end{ack}

\clearpage
\bibliographystyle{plainnat}
\bibliography{reference}

\clearpage
\appendix

\section{Formal Connection to Iterative Amortized Inference}
\label{sec:appendix-formal}

We formalize the connection between LEE's iterative search and the framework of \citet{marino2018iterative}. Table~\ref{tab:search-comparison} summarizes the high-level differences against GenSR's CMA-ES; the rest of this section makes the LEE--Marino correspondence precise.

\begin{table}[h]
\caption{\textbf{Search mechanism comparison.} LEE's iterative update uses the model's own encoder as the optimizer; each step conditions on the full candidate expression and the data, whereas CMA-ES sees only a scalar fitness.}
\label{tab:search-comparison}
\centering
\small
\begin{tabular}{lcc}
\toprule
& \textbf{GenSR (CMA-ES)} & \textbf{LEE (Iterative Amortized)} \\
\midrule
Update rule & $\bm{z}_{t+1} \leftarrow \text{CMA-ES}(\bm{z}_t, \text{fitness})$ & $\bm{z}_{t+1} \leftarrow f_\theta(g_\text{expr}(\bm{z}_t), \mathcal{D})$ \\
Signal per step & Scalar fitness ($R^2$) & Full token sequence + scatter \\
Model awareness & Black-box (ignores encoder) & Uses encoder's learned geometry \\
Cost scaling & $O(d_z^2)$ covariance updates & One encoder forward per step \\
Latent codes & Two separate ($z_\text{sym}$, $z_\text{num}$) & One shared $\bm{z}$ \\
\bottomrule
\end{tabular}
\end{table}

\noindent\textbf{Setup.}
Following the notation of \citet{marino2018iterative}, let $\bm{\lambda}^{(i)}$ denote the approximate posterior parameters for data example $\bm{x}^{(i)}$, and let $\mathcal{L}(\bm{x}^{(i)}, \bm{\lambda}^{(i)}; \theta)$ be the ELBO.
Standard amortized inference uses a direct mapping (their Eq.~5):
\begin{equation}
\bm{\lambda}^{(i)} \leftarrow f(\bm{x}^{(i)}; \phi).
\tag{A.1}
\end{equation}
Iterative amortized inference refines this estimate (their Eq.~6):
\begin{equation}
\bm{\lambda}_{t+1}^{(i)} \leftarrow f_t\bigl(\nabla_{\bm{\lambda}} \mathcal{L}_t^{(i)},\; \bm{\lambda}_t^{(i)};\; \phi\bigr).
\tag{A.2}
\end{equation}

\noindent\textbf{LEE as an instance.}
In LEE, the approximate posterior is parameterized by the latent code $\bm{z} \in \mathbb{R}^{d_z}$ (i.e., $\bm{\lambda} \equiv \bm{z}$), from which expressions are decoded autoregressively.
The ELBO analog is:
\begin{equation}
\mathcal{L}(\mathcal{D}, \bm{z}; \theta) = \underbrace{\log p_\theta(\bm{t} \mid \bm{z})}_{\text{expression fit}} + \underbrace{\log p_\theta(\bm{y} \mid \bm{z}, \bm{X})}_{\text{evaluation fit}} - \underbrace{D_\text{KL}(q(\bm{z}) \| p(\bm{z}))}_{\text{regularization}}.
\tag{A.3}
\end{equation}
LEE's iterative update replaces the gradient $\nabla_{\bm{z}} \mathcal{L}$ with the decoded expression $\hat{e}_t = g_\text{expr}(\bm{z}_t)$:
\begin{equation}
\bm{z}_{t+1} = f_\theta\bigl(\hat{e}_t,\; \mathcal{D}\bigr) = f_\theta\bigl(g_\text{expr}(\bm{z}_t),\; \mathcal{D};\; \theta\bigr).
\tag{A.4}
\end{equation}
This corresponds to the \emph{error-encoding} variant (their Eq.~14), where the bottom-up mismatch between $\hat{e}_t(\bm{X})$ and $\bm{y}$ and the top-down discrepancy between $\hat{e}_t$'s structure and the latent prior are implicitly computed by the encoder's cross-attention.

\noindent\textbf{Key difference.}
In \citet{marino2018iterative}, the error signal is a real-valued vector in the same space as $\bm{\lambda}$.
In LEE, the ``error signal'' is the \emph{symbolic expression} $\hat{e}_t$---a discrete, structured object.
The encoder $f_\theta$ performs the nontrivial mapping from this structured input to a continuous update in $\mathcal{Z}$, which is why training with $\mathcal{L}_\text{refine}$ (Eq.~\ref{eq:refine}) is essential: it teaches the encoder how to extract useful refinement signals from imperfect expressions.

\section{Mechanism Analysis}
\label{sec:appendix-analysis}

\noindent\textbf{Why does re-encoding close the amortization gap?}
The encoder computes fundamentally different functions in scatter-only vs.\ joint mode. In scatter-only mode ($t{=}0$), it solves an ambiguous inverse problem: map finite, noisy observations to a latent code. In joint mode ($t{>}0$) it receives both a candidate expression $\hat{e}_t$ and $\mathcal{D}$, and can internally attend to their mismatch---effectively computing a residual:
\begin{equation}
\bm{z}_{t+1} = f_\theta\bigl(\hat{e}_t, \mathcal{D}\bigr) \approx f_\theta\bigl(\hat{e}_t,\; \hat{e}_t(\bm{X}) - \bm{y},\; \bm{X}\bigr),
\label{eq:error-encoding}
\end{equation}
where $\hat{e}_t(\bm{X}) - \bm{y}$ is the residual the encoder computes internally via cross-attention between the symbolic and numeric streams. This mirrors the ``error encoding'' variant of \citet{marino2018iterative} (their Eq.~14), which was shown to approximate higher-order derivatives and converge faster than gradient encoding. The refinement loss $\mathcal{L}_\text{refine}$ (Eq.~\ref{eq:refine}) explicitly trains the encoder for this regime: at inference, decoded expressions from the pool play the role of the noisy input $\tilde e$ that the encoder has been trained to denoise.

\noindent\textbf{Expression simplicity as inductive bias.}
LEE's tendency toward simple expressions arises from three compounding effects: (1)~the autoregressive decoder has an implicit length bias---shorter token sequences have higher probability under teacher forcing; (2)~the pool scoring function $s(e) = R^2(e) - \alpha \cdot C(e)$ explicitly favors parsimony; and (3)~the evaluation decoder creates a latent geometry where simple functional forms---more prevalent in the training distribution---occupy larger volumes of $\mathcal{Z}$ and are therefore more likely to be decoded. GenSR and GP methods, in contrast, have no inherent simplicity bias and rely on post-hoc complexity penalties.

\section{Iterative Convergence}
\label{sec:appendix-convergence}

\begin{figure}[h]
\centering
\begin{subfigure}[b]{0.48\textwidth}
\centering
\includegraphics[width=\linewidth]{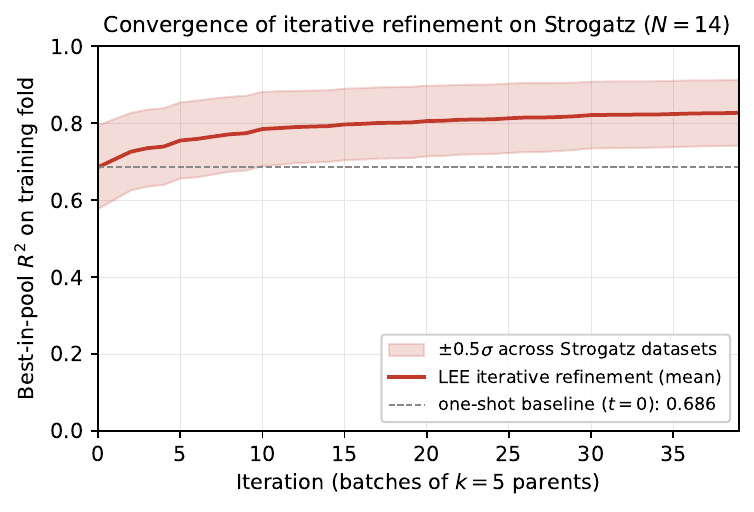}
\caption{$R^2$ convergence over iterations}
\label{fig:convergence-r2}
\end{subfigure}
\hfill
\begin{subfigure}[b]{0.48\textwidth}
\centering
\includegraphics[width=\linewidth]{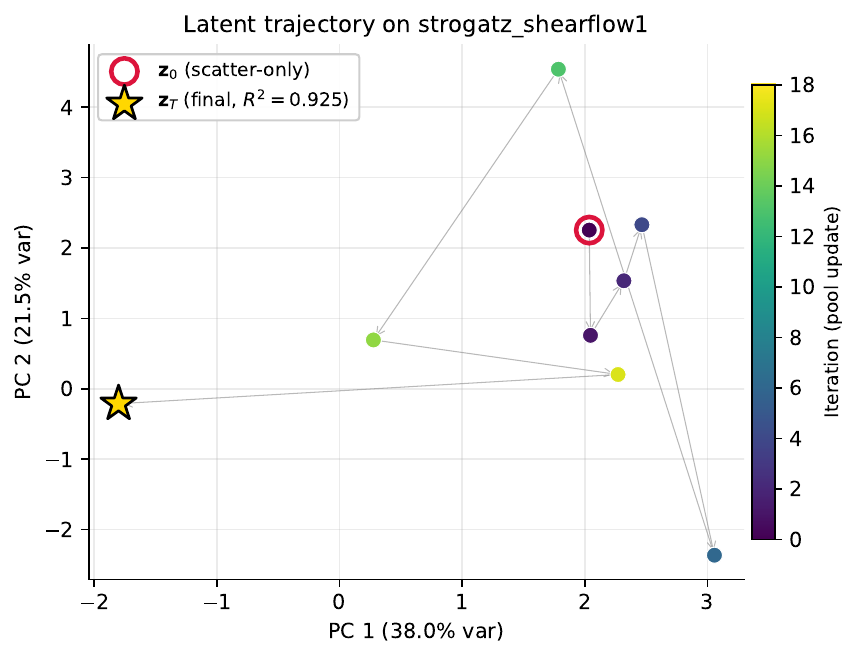}
\caption{Latent trajectory (\texttt{strogatz\_shearflow1})}
\label{fig:trajectory}
\end{subfigure}
\caption{\textbf{Iterative convergence.} (a)~Best-in-pool $R^2$ (mean $\pm$ $0.5\sigma$) across the 14 Strogatz datasets. (b)~PCA projection of the latent vectors $\bm{z}$ produced at successive pool-champion updates.}
\label{fig:convergence}
\end{figure}

Figure~\ref{fig:convergence} summarizes the convergence behavior of LEE's iterative refinement on the 14 Strogatz datasets. The pool's best $R^2$ (panel a) improves rapidly in the first $10$--$15$ batches as re-encoding corrects the initial scatter-only estimate, then refines gradually toward a within-round pool-best of $0.827$ (up from a $t{=}0$ baseline of $0.686$); the full best-of-$R$ pipeline reaches $0.854$ in Table~\ref{tab:main}. Panel (b) visualizes the same refinement in the latent space: successive pool-champion encodings on \texttt{strogatz\_shearflow1} trace a path from $\bm{z}_0$ (scatter-only initialization) to $\bm{z}_T$ ($R^2{=}0.925$), so each re-encoding step translates a discrete improvement in the decoded expression into a measurable move in $\mathcal{Z}$.

\section{Latent Space Interpolation}
\label{sec:appendix-interpolation}

A complementary qualitative test of the latent geometry is whether linear interpolation between two encoded equations decodes to expressions that smoothly bridge them in function space. We encode two ground-truth expressions $A$ and $B$ jointly with their scatter observations to obtain $\bm z_A = f_\theta(\text{tokens}_A, \mathcal D_A)$ and $\bm z_B = f_\theta(\text{tokens}_B, \mathcal D_B)$; for each $t \in \{0, 1/3, 2/3, 1\}$ we set $\bm z_t = (1{-}t)\bm z_A + t\bm z_B$, decode several candidates ($1$ greedy + $31$ samples at $\tau{=}0.7$), and pick the one whose values most closely match the linear blend $(1{-}t)\,y_A + t\,y_B$.

\begin{figure}[h]
\centering
\includegraphics[width=0.62\textwidth]{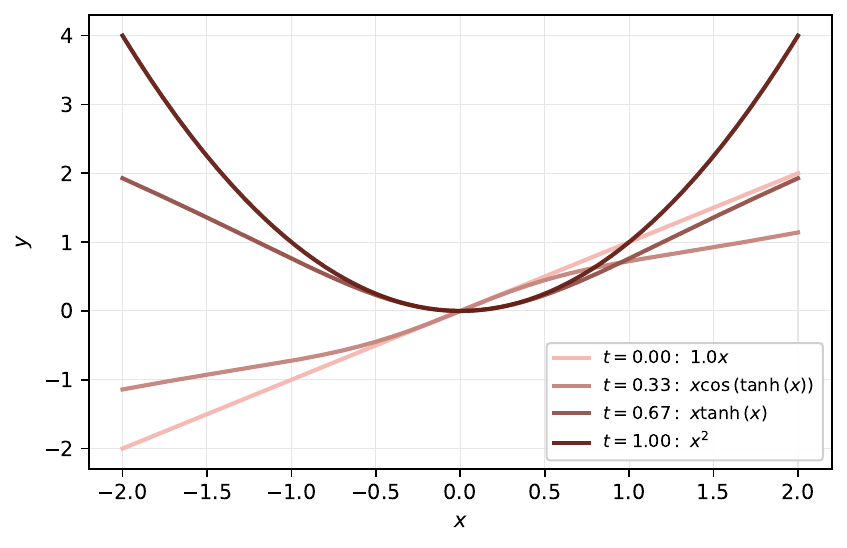}
\caption{\textbf{Latent interpolation, $y{=}x \to y{=}x^2$ on $x\in[-2,2]$.} Solid curves are decoded expressions at four points $\bm z_t = (1{-}t)\bm z_A + t\bm z_B$; the color gradient (light $\to$ dark) encodes $t$. Endpoints round-trip back to their inputs ($t{=}0$ decodes to $x$, $t{=}1$ to $x^2$); intermediate latents at $t{=}1/3$ and $t{=}2/3$ decode to syntactically distinct but functionally smooth interpolants ($x\cos(\tanh x)$ and $x\tanh x$).}
\label{fig:interp}
\end{figure}

Figure~\ref{fig:interp} shows the result for $A{:}\ y=x \to B{:}\ y=x^2$. The endpoints round-trip back to their inputs, and the intermediate decoded expressions trace a smooth deformation from a linear ramp through a $\tanh$-modulated bowl into the parabola. This qualitative behavior is consistent with the functional-grounding hypothesis (Sec.~\ref{sec:eval-decoder}): the evaluation decoder shapes $\mathcal{Z}$ so that proximity in latent space reflects functional similarity.

\section{VAE Architecture Ablation}
\label{sec:appendix-vae-ablation}

To probe the contribution of the VAE encoder with conditional-KL alignment (Sec.~\ref{sec:training}), we compare the full LEE model against a non-VAE variant trained on the same data: deterministic encoder, $z$-norm penalty in place of KL, no cross-modal alignment loss. We compare on two axes: (i) headline accuracy on Strogatz at $\epsilon{=}0.1$ (Table~\ref{tab:vae-ablation}), and (ii) the four-step latent interpolation introduced in Appendix~\ref{sec:appendix-interpolation} (Figure~\ref{fig:vae-ablation}). Both probes use the same evaluation protocol and inference hyperparameters; only the model checkpoint differs.

\begin{table}[h]
\caption{\textbf{VAE architecture ablation on Strogatz $\epsilon{=}0.1$} (mean $\pm$ std over $3$ trials with failed-to-converge datasets dropped from the $R^2$ and complexity averages). Removing the VAE encoder + KL alignment costs ${\sim}0.30$ $R^2$ on Strogatz at $\epsilon{=}0.1$; on $1$--$2$ of $14$ datasets per trial the non-VAE variant fails to return a valid expression at all.}
\label{tab:vae-ablation}
\centering
\small
\begin{tabular}{lccc}
\toprule
Variant & $R^2$ ($\uparrow$) & Cmplx ($\downarrow$) & Failures / 14 \\
\midrule
Non-VAE variant ($z$-norm + no KL align)     & $0.582 \pm 0.084$ & $8.5 \pm 0.9$ & $1.3 \pm 0.6$ \\
\textbf{Full LEE (VAE + conditional KL)}     & $\mathbf{0.880 \pm 0.024}$ & $\mathbf{8.3 \pm 1.4}$ & $0$ \\
\bottomrule
\end{tabular}
\end{table}

\begin{figure}[h]
\centering
\includegraphics[width=\textwidth]{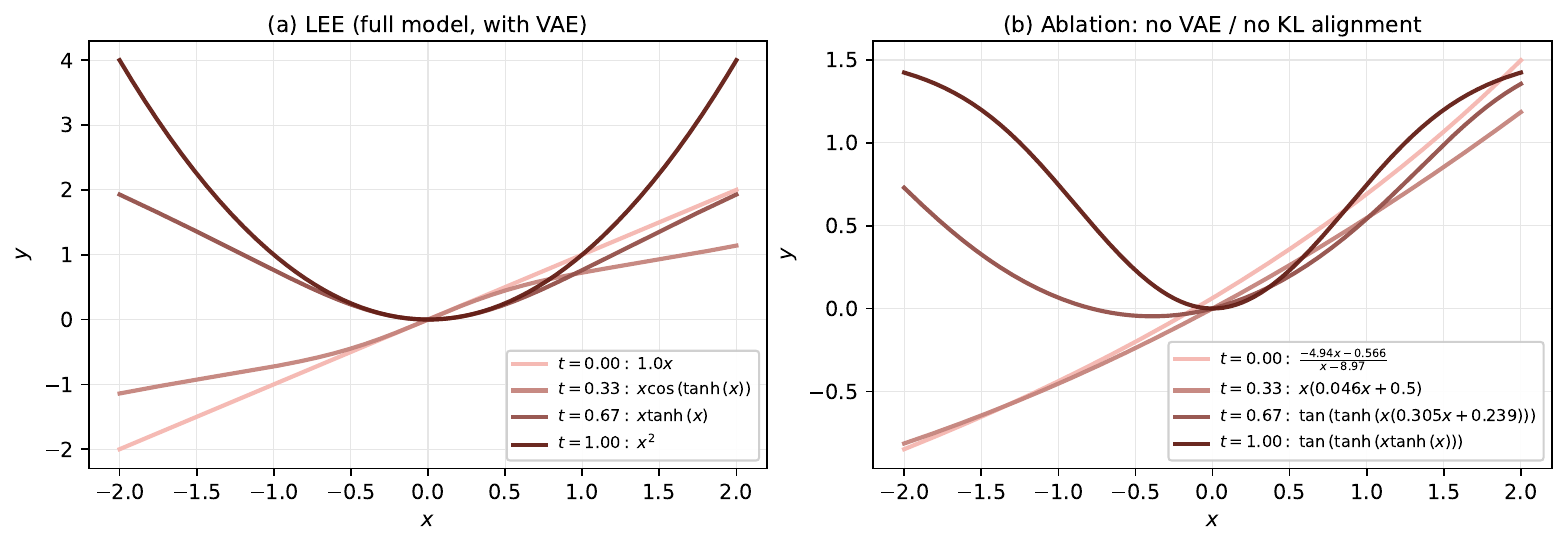}
\caption{\textbf{Latent interpolation comparison: with vs.\ without VAE} ($y{=}x \to y{=}x^2$ on $x\in[-2,2]$). Solid curves are decoded expressions at $t \in \{0, 1/3, 2/3, 1\}$; the color gradient encodes $t$. \textbf{(a)} Full LEE round-trips both endpoints and the intermediates are syntactically distinct but functionally smooth ($x\cos(\tanh x)$ and $x\tanh x$). \textbf{(b)} The non-VAE variant fails to round-trip ($t{=}1$ decodes to a saturating bowl rather than $x^2$) and intermediates are ill-conditioned $\tan(\tanh(\cdot))$ shapes that bear no resemblance to either $A$ or $B$.}
\label{fig:vae-ablation}
\end{figure}

The accuracy gap (Table~\ref{tab:vae-ablation}) and the qualitative interpolation breakdown (Figure~\ref{fig:vae-ablation}) tell the same story: the VAE-induced latent regularity is essential for a step in $\mathcal{Z}$ to correspond to a step in function space, which is the geometric prerequisite for the iterative re-encoding update of Eq.~\ref{eq:lee-update} to work as designed.

\textit{Caveat:} this non-VAE variant differs from the full LEE not only in lacking the VAE/KL terms but also in the cross-modal alignment loss and in being trained on a smaller variable vocabulary (5 vs.\ 10), so the comparison is a proxy for the VAE ablation rather than an exact controlled study; a from-scratch retrain with $\texttt{use\_vae}{=}\texttt{false}$ holding all other components fixed is left for future work.

\section{Latent Space Property Decoding}
\label{sec:appendix-property-probe}

To test whether the encoder's latent space encodes structurally meaningful properties of the input expression, we ask: for each of seven properties of an expression $e$, is the property linearly recoverable from $\bm z = f_\theta(e, \mathcal{D})$ alone---i.e., is there a single direction in $\mathcal{Z}$ along which expressions satisfying the property are separated from those that do not?

\noindent\textbf{Setup.}
We sample $5000$ expressions uniformly at random from the test split of our pre-training corpus, encode each through the full encoder on freshly generated scatter $\mathcal{D}$ (yielding a $5000{\times}512$ latent matrix), and assign each example labels for the seven properties listed in Table~\ref{tab:property-probe}. For each property, we train a logistic regression classifier (a single linear vector with $L_2$ penalty, $C{=}1$) on a stratified $4000$/$1000$ train/test split and report test accuracy together with the test AUC for binary targets. The 10-way \emph{num\_variables} classifier is multinomial and is reported as accuracy only.

\begin{table}[h]
\caption{\textbf{Property linear-probe panel.} Test accuracy and AUC of a single-vector logistic regression on $\bm z \in \mathbb{R}^{512}$. The \emph{True \%} column reports the proportion of positives in the sampled subset (chance for the 10-class probe is $10\%$). All binary probes achieve AUC $\ge 0.86$, indicating each property has a dedicated linear direction in $\mathcal{Z}$.}
\label{tab:property-probe}
\centering
\small
\begin{tabular}{lrrr}
\toprule
Property & True \% & Test acc & Test AUC \\
\midrule
has trig ($\sin/\cos/\tan/\tanh$)        & 53.7 & 0.831 & 0.914 \\
has log/exp                              & 26.8 & 0.865 & 0.912 \\
has sq/cube ($x^2$ or $x^3$)             & 33.5 & 0.831 & 0.861 \\
has division                             & 73.7 & 0.886 & 0.944 \\
is polynomial (no trig, log/exp, abs, $\sqrt{\cdot}$) & 26.5 & 0.866 & 0.938 \\
high-dim ($n_\text{vars} \ge 5$)         & 31.6 & 0.939 & 0.982 \\
num\_variables (10-class)                & 10.0 (chance) & 0.727 & --- \\
\bottomrule
\end{tabular}
\end{table}

\noindent\textbf{Findings.}
All six binary properties are linearly recoverable with AUC between $0.86$ and $0.98$. The strongest axis is \emph{high-dim} (AUC $0.98$, accuracy $94\%$), followed by \emph{has division} (AUC $0.94$) and \emph{is polynomial} (AUC $0.94$); the weakest is \emph{has sq/cube} (AUC $0.86$), still well above chance. The 10-way variable-count classifier reaches $72.7\%$ accuracy ($7.3\times$ chance), comparable to the AUC profile of the binary probes. Each property, therefore, corresponds to a distinct linear direction in $\mathcal{Z}$, and the directions evidently coexist: a $512$-dimensional space has ample capacity to host one axis per property without conflict. This supports the design intent of Sec.~\ref{sec:eval-decoder}---the encoder learns a function-grounded latent space whose principal directions correspond to interpretable structural properties of the underlying expression---and helps explain why a small number of iterative re-encoding steps suffices to traverse the space (Sec.~\ref{sec:iterative}): movement along any single axis carries a structurally meaningful change in the decoded expression.

\section{Training Details}
\label{sec:appendix-training}

\noindent\textbf{Tokenization.}
Expressions are serialized in prefix (Polish) notation over a vocabulary of 40 tokens: 4 special tokens (\texttt{PAD}, \texttt{BOS}, \texttt{EOS}, \texttt{UNK}), 2 structural tokens, 10 variables ($x_0, \ldots, x_9$), 15 operators ($+, -, \times, \div, \sin, \cos, \tan, \tanh, \exp, \log, \sqrt{}, x^2, x^3, \text{abs}, \text{neg}$), and 14 digit tokens for constant encoding.
Constants are represented at 3 significant figures in scientific notation as 9-token sequences: $[\text{sign}, d_1, \texttt{.}, d_2, d_3, \texttt{e}, \text{sign}', e_1, e_2]$.

\noindent\textbf{Grammar and expression sampling.}
The stochastic context-free grammar (SCFG) follows the protocol of \citet{lample2019deep,kamienny2022end}: a binary-tree scaffold is sampled first (with $1\!\leq\!b\!\leq\!b_\text{max}{=}4$ binary operators), then unary operators ($u$ drawn with $u_\text{max}{=}4$) are attached, and finally leaf nodes are filled with variables (uniformly over the allowed set) and numerical constants. Constants are drawn from a mixture: $60\%$ integer $[-10,10]$, $30\%$ log-uniform over $[10^{-2}, 10^{2}]$, and $10\%$ from a small catalogue of physics constants ($\pi$, $e$, etc.). We enforce \emph{variable coverage}: every declared variable appears at least once in the tree. Trees are re-sampled on syntactic failure (NaN/Inf on the fixed query grid). The final training corpus contains ${\sim}13.4$M unique prefix sequences, split 80/10/10 into train/val/test.

\noindent\textbf{Pre-training cost.}
Total training wallclock is $\approx 200$ GH200-GPU-hours spread across the five phases of Table~\ref{tab:loss-weights}, on a single node with batch size 256 and mixed-precision (bf16) forward/backward. The dataset is generated offline in ${\sim}8$ CPU-hours on 16 cores. This up-front cost is amortized across downstream datasets; for SR workflows that evaluate hundreds of datasets, the break-even point against 1--24-hour-per-dataset GP baselines is in the single digits.

\begin{table}[h]
\caption{\textbf{Loss weights across the five training phases.}}
\label{tab:loss-weights}
\centering
\small
\begin{tabular}{lccccc}
\toprule
Phase & $\lambda_\text{expr}$ & $\lambda_\text{eval}$ & $\lambda_\text{KL}$ & $\lambda_\text{align}$ & $\lambda_\text{refine}$ \\
\midrule
1 (basic) & 1.0 & 5.0 & 0.001 & 0 & 0 \\
2 (+align) & 1.0 & 5.0 & 0.001 & 2.0 & 0 \\
3 (+refine) & 1.0 & 5.0 & 0.001 & 2.0 & 1.0 \\
4 (freeze dec) & 0 & 0 & 0.001 & 5.0 & 0 \\
5 (unfreeze) & 1.0 & 5.0 & 0.001 & 2.0 & 1.0 \\
\bottomrule
\end{tabular}
\end{table}

\noindent\textbf{Training schedule.}
Phase~1: 50k steps (basic reconstruction, all parameters).
Phase~2: 30k steps (add alignment, encoder focused).
Phase~3: 50k steps (add refinement, full model).
Phase~4: 30k steps (freeze decoders, alignment-only, encoder learns modality bridging).
Phase~5: 40k steps (unfreeze all, co-adaptation).
Batch size 256, AdamW ($\beta_1{=}0.9$, $\beta_2{=}0.999$), cosine decay from $3{\times}10^{-4}$ to $1{\times}10^{-5}$.

\noindent\textbf{Data augmentation.}
During training, scatter points are randomly sub-sampled (128--200 points per example) and coordinate-rotated for multi-variable expressions.
Token corruption for $\mathcal{L}_\text{refine}$: each token is independently dropped (15\%), swapped with a random token (10\%), or kept (75\%).

\noindent\textbf{Constant optimization.}
After decoding, numerical constants in each expression are refined by L-BFGS-B, minimizing MSE on the training split.
The budget ramps linearly from 100 to 300 steps over the search iterations.
For datasets with $>1000$ training points, we randomly subsample 1000 points for each L-BFGS-B call.

\section{Architecture Details}
\label{sec:appendix-arch}

\begin{table}[h]
\caption{\textbf{Architecture hyperparameters.}}
\label{tab:arch}
\centering
\small
\begin{tabular}{lccc}
\toprule
& Encoder & Expr.\ Decoder & Eval.\ Decoder \\
\midrule
Model dim ($d$) & 768 & 512 & 512 \\
Layers & 6 & 8 & 4 \\
Heads & 12 & 8 & 8 \\
FFN dim & 3072 & 2048 & 2048 \\
Dropout & 0.1 & 0.1 & 0.1 \\
Memory tokens ($K$) & --- & 4 & 4 \\
Latent dim ($d_z$) & 512 & 512 & 512 \\
\midrule
Parameters & ${\approx}$75M & ${\approx}$50M & ${\approx}$25M \\
\bottomrule
\end{tabular}
\end{table}

The encoder's scatter-embedding MLP has a hidden dimension of 256 and uses SiLU activation. The log-compressed coordinate $\tilde x = \text{sign}(x)\log(1+|x|)$ is additionally divided by a fixed scale of $4$ before the MLP, so that $|x|\!\leq\!50$ lands roughly in $[-1,1]$; function values are not rescaled.
The evaluation decoder's query embedding MLP maps $k$ coordinate dimensions to $d_\text{eval}$, with a hidden dimension of 256.
Both decoders project $\bm{z}$ into $K{=}4$ memory tokens via a linear layer, then use the Transformer decoder cross-attention to these memory tokens.
The expression decoder uses causal self-attention; the evaluation decoder uses bidirectional self-attention (queries can attend to all other query positions).

\section{Inference Procedures}
\label{sec:appendix-inference}

\noindent\textbf{Full inference hyperparameters.}
Table~\ref{tab:inference-hparams} lists every inference-side hyperparameter and its default value. All numbers reported in Sec.~\ref{sec:results} use these settings unless noted otherwise.

\begin{table}[h]
\caption{\textbf{Inference hyperparameters.}}
\label{tab:inference-hparams}
\centering
\small
\begin{tabular}{lcl}
\toprule
Name & Value & Meaning \\
\midrule
$R$ & 10 & rounds per trial \\
$T$ & 200 & refinement iterations per round \\
$P$ & 16 & candidate pool size \\
$n_\text{init}$ & 32 & candidates decoded from $\bm z_0$ at round start \\
$n_\text{new}$ & 3 & new candidates decoded per iteration \\
batch $k$ & 5 & parents processed per batch \\
refresh period & 5 & batches between scatter resamples \\
$\alpha$ & 0.002 & complexity penalty in scoring $R^2_\text{train}-\alpha C(e)$ \\
$\tau$ & 0.7 & decoder sampling temperature \\
L-BFGS-B budget & 100 $\to$ 300 & constant-refinement steps, linearly ramped over $R$ \\
L-BFGS-B subsample & 1000 & rows used for constant fit if $N_\text{train}{>}1000$ \\
$n_\text{grad}$ & 50 & gradient refinement steps per segment \\
$\eta$ & $5\!\times\!10^{-3}$ & gradient refinement learning rate \\
$\lambda_\text{prox}$ & 0.1 & proximal anchor weight \\
decode period & 25 & iterations between gradient segments \\
MAX\_SEARCH\_POINTS & 2000 & row cap for per-candidate $R^2$ scoring \\
$n_\text{trials}$ & 10 & independent trials per (dataset, $\epsilon$) for main results (3 for ablations) \\
\bottomrule
\end{tabular}
\end{table}

\noindent\textbf{Pool initialization.}
The initial latent $\bm z_0 = f_\theta(\texttt{pad}, \mathcal{D})$ is decoded into $n_\text{init}=32$ candidates: 1 greedy argmax decode and 31 temperature-$\tau$ sampled decodes. Each candidate's constants are immediately refined via L-BFGS-B on the training fold. Candidates are scored by $s(e) = \text{clip}(R^2_\text{train}(e), -1, 1) - \alpha\,C(e)$, with $C(e)$ the SymPy-simplified node count. The top $P=16$ form the initial pool $\Pi_0$.

\noindent\textbf{Parent selection and complexity diversity.}
At each iteration we sample $k=5$ parents from $\Pi_t$ with rank-weighted probabilities $p_i \propto 1/(i+1)$ (so rank-1 is twice as likely as rank-3). When merging new candidates back, we enforce \emph{complexity diversity} by bucketing $C(e)$ into 4 bins $[0, 5)$, $[5, 10)$, $[10, 20)$, $[20, \infty)$ and capping the pool at $\lceil P/4\rceil=4$ entries per bucket (best by $s(e)$). This prevents the pool from collapsing onto a single expression family.

\noindent\textbf{Scatter refresh.}
Every 5 batches, we (i) re-sample $n_\text{scatter}=200$ scatter points from the training fold, (ii) re-compute a fresh scatter-only $\bm z_0$, and (iii) decode 3 new candidates that are added to the pool before the next iteration. This injects exploration when the pool has converged.

\noindent\textbf{Gradient refinement (pg mode).}
Every 25 iterations of iterative refinement, we take the current pool champion $\bm z^*$ (its constants fixed), run 50 Adam steps on $\bm z$ minimizing $\|g_\text{eval}(\bm z, \bm X_\text{train}) - \bm y_\text{train}\|_2^2 + \lambda_\text{prox}\|\bm z - \bm z^*\|_2^2$, decode from the resulting $\bm z$, and insert the decoded expression (after L-BFGS-B) back into the pool. The proximal term with $\lambda_\text{prox}=0.1$ prevents the gradient descent from wandering off the decodable manifold.

\noindent\textbf{Safety fallback.}
After pg-mode concludes a round, the reported winner is the pool entry with the highest validation $R^2$; if that winner was produced by a gradient segment but has \emph{lower} validation $R^2$ than the best pool entry from the previous iterative-only step, we revert to the latter. This is the safety fallback of Sec.~\ref{sec:iterative}(c) and ensures the combined mode is never worse than iterative-only up to selection noise.

\noindent\textbf{Seeding and reproducibility.}
Each trial uses a fresh random seed $s_t = s_\text{base} + 1000\,t$, where $s_\text{base}$ varies per trial $t \in \{0, \ldots, n_\text{trials}-1\}$ (with $n_\text{trials}{=}10$ for the main results in Table~\ref{tab:main} and $n_\text{trials}{=}3$ for the ablations). Within a trial, $s_t$ controls (i) the 60/15/25 train/val/test split, (ii) all decoder sampling, (iii) gradient optimizer initialization, and (iv) scatter subsampling. All trials share the same pre-trained checkpoint. Baseline numbers follow the seed protocol of their source publication.

\noindent\textbf{Pool size sensitivity.}
We sweep the candidate pool size $P \in \{8, 16, 32\}$ on Strogatz at $\epsilon{=}0.1$, holding every other inference hyperparameter fixed (Table~\ref{tab:pool-sweep}). The default $P{=}16$ used in the main results sits at a small but consistent sweet spot: $P{=}8$ slightly underperforms because the rank-weighted parent sampler depletes diversity too quickly, while $P{=}32$ slightly underperforms because lower-rank parents are sampled too rarely to inject fresh exploration.

\begin{table}[h]
\caption{\textbf{Pool size sensitivity} (Strogatz, $\epsilon{=}0.1$, $3$ trials, paper protocol). The default $P{=}16$ used in the main results sits at a small but consistent sweet spot between under- and over-sized pools.}
\label{tab:pool-sweep}
\centering
\small
\begin{tabular}{ccc}
\toprule
$P$ & $R^2$ ($\uparrow$) & Cmplx ($\downarrow$) \\
\midrule
$8$    & $0.863 \pm 0.030$ & $10.5 \pm 0.5$ \\
$\mathbf{16}$ (default) & $\mathbf{0.880 \pm 0.024}$ & $\mathbf{8.3 \pm 1.4}$ \\
$32$   & $0.863 \pm 0.026$ & $10.7 \pm 0.9$ \\
\bottomrule
\end{tabular}
\end{table}

\section{Timing Details}
\label{app:timing}

We move detailed wall clock comparisons here because hardware across methods varies substantially, and absolute seconds are therefore not directly comparable. What follows are the operational timings of our method and the reported timings of the baselines from the published SRBench and GenSR data.

\noindent\textbf{LEE: per-dataset wallclock.}
One full LEE run on a dataset consists of $R=10$ rounds of $T=200$ iterations of iterative\,+\,gradient refinement, executed with $W=8$ parallel workers on a single NVIDIA GH200. Per-dataset wallclock is therefore lower-bounded by $\lceil R/W\rceil\cdot\tau = 2\tau$ (where $\tau$ is the mean per-round time), with measured values closer to $1.25\tau$ under continuous round dispatch. Table~\ref{tab:timing} reports the measured wallclock averaged over all datasets in each benchmark group and $10$ trials.

\begin{table}[h]
\caption{\textbf{LEE per-dataset wallclock time} (seconds, mean over $10$ trials, single GH200 node with 16 CPU cores, 8-way round parallelism). Values are the full cost to produce the best-of-10-rounds result for a single dataset.}
\label{tab:timing}
\centering
\small
\begin{tabular}{lccc}
\toprule
Benchmark & $\epsilon{=}0$ & $\epsilon{=}0.01$ & $\epsilon{=}0.1$ \\
\midrule
Strogatz (14)   & 64.8 $\pm$ 1.3  & 65.6 $\pm$ 1.1  & 67.6 $\pm$ 1.0  \\
Feynman (116)   & 109.2 $\pm$ 1.0 & 109.8 $\pm$ 1.2 & 116.8 $\pm$ 0.3 \\
Black-box (63)  & 79.2 $\pm$ 1.1  & --- & --- \\
\midrule
Strogatz, iterative only & 48.2 & --- & 49.1 \\
Strogatz, gradient only  &  5.0 & --- &  5.3 \\
\bottomrule
\end{tabular}
\end{table}

\noindent\textbf{Comparison to baselines.}
Reported times for baselines in Table~\ref{tab:main} of the main text vary from \textasciitilde 4 seconds (E2ESR one-shot on Feynman) to 149k seconds (SBP-GP on black-box). LEE's per-dataset cost is in the tens-of-seconds regime: an order of magnitude faster than most GP methods (Operon, GP-GOMEA, SBP-GP) and within $\sim\!30\times$ of the fastest one-shot neural methods (E2ESR, SNIP, NeurSR), which perform no iterative search at all. In absolute terms, a full pass over the 116 Feynman datasets at $\epsilon{=}0.1$ completes in roughly $116 \times 117/60 \approx 3.3$ hours of wallclock on a single node.

\noindent\textbf{Where the time goes.}
Within each round, L-BFGS-B constant refinement dominates (roughly 65--75\% of $\tau$), followed by GPU forward passes through encoder/decoders (20--30\%) and expression serialization/SymPy simplification (the remainder). The gradient segments in the combined mode are inexpensive ($<\!5\%$ of $\tau$), which is consistent with gradient refinement alone being the fastest mode in Table~\ref{tab:timing} ($\sim$5 s) but the weakest in accuracy: most of the wallclock cost is in evaluating and refining candidate expressions, which only iterative refinement exercises.

\noindent\textbf{Caveats.}
Baseline times were measured on the hardware reported in their respective source papers; some (e.g.\ SBP-GP) use server-class CPUs over days, others use single-GPU workstations. Our numbers assume a single-node GH200 + 16 CPU cores. We therefore \emph{intentionally} omit time from the main-text Table~\ref{tab:main} and use it here only as a coarse order-of-magnitude reference.

\section{LEE Summary Statistics with Standard Deviations}
\label{sec:appendix-std}

Table~\ref{tab:lee-std} reports the per-group LEE mean $\pm$ standard deviation for $R^2$ and complexity, omitted from the main Table~\ref{tab:main} for compactness. Each row aggregates $10$ independent trials with distinct splits and search seeds (Sec.~\ref{sec:experiments}); the std reflects trial-to-trial variation.

\begin{table}[h]
\caption{\textbf{LEE per-group summary with standard deviations} (mean $\pm$ std over $10$ trials; same trials underlying Table~\ref{tab:main}).}
\label{tab:lee-std}
\centering
\small
\begin{tabular}{lcc}
\toprule
Setting & $R^2$ ($\uparrow$) & Cmplx ($\downarrow$) \\
\midrule
\textit{Strogatz (14)} & & \\
\quad $\epsilon{=}0$    & $0.854 \pm 0.005$ & $8.1 \pm 1.5$ \\
\quad $\epsilon{=}0.01$ & $0.876 \pm 0.012$ & $8.9 \pm 1.2$ \\
\quad $\epsilon{=}0.1$  & $0.880 \pm 0.024$ & $8.3 \pm 1.4$ \\
\midrule
\textit{Feynman (116)} & & \\
\quad $\epsilon{=}0$    & $0.884 \pm 0.001$ & $9.9 \pm 0.5$ \\
\quad $\epsilon{=}0.01$ & $0.884 \pm 0.005$ & $10.1 \pm 0.3$ \\
\quad $\epsilon{=}0.1$  & $0.824 \pm 0.005$ & $10.6 \pm 0.7$ \\
\midrule
\textit{Black-box (63)} & $0.559 \pm 0.004$ & $9.0 \pm 0.4$ \\
\bottomrule
\end{tabular}
\end{table}

Standard deviations are small across the board: $R^2$ std ranges from $0.001$ to $0.024$, with the largest variance on noisy Strogatz ($\epsilon{=}0.1$, $\sigma{=}0.024$), reflecting the higher sensitivity of small-dataset noise realizations. Complexity std is similarly tight (${\le}1.5$), confirming that the simplicity property of LEE is consistent across trials, not an artifact of a lucky seed.

\section{Additional Pareto Frontiers}
\label{sec:appendix-pareto}

\begin{figure}[h]
\centering
\begin{subfigure}[b]{0.48\textwidth}\includegraphics[width=\textwidth]{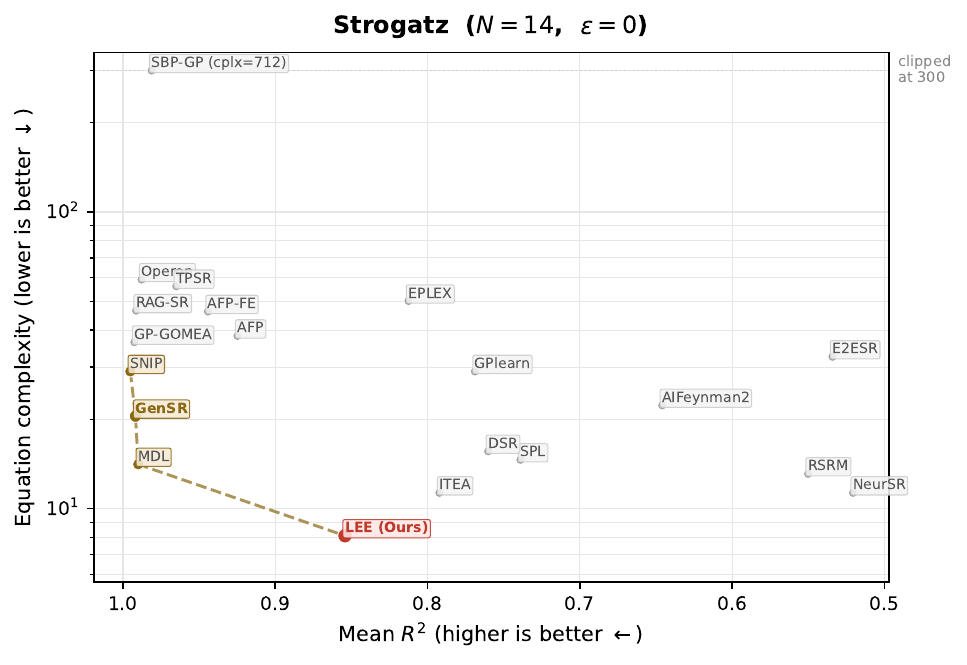}
\caption{Strogatz, $\epsilon{=}0$}\end{subfigure}\hfill
\begin{subfigure}[b]{0.48\textwidth}\includegraphics[width=\textwidth]{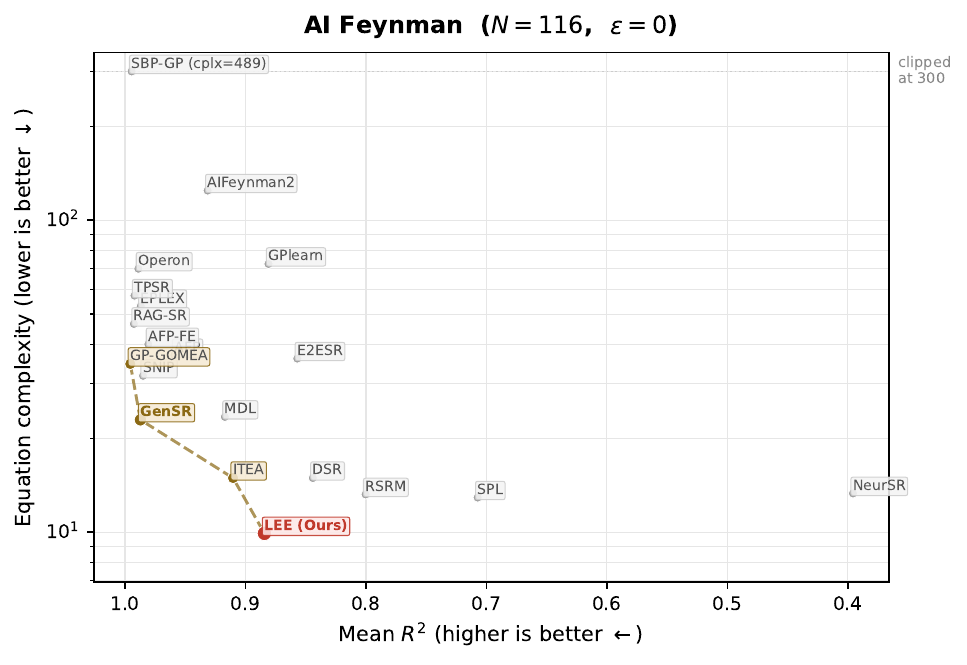}
\caption{Feynman, $\epsilon{=}0$}\end{subfigure}

\vspace{4pt}
\begin{subfigure}[b]{0.48\textwidth}\includegraphics[width=\textwidth]{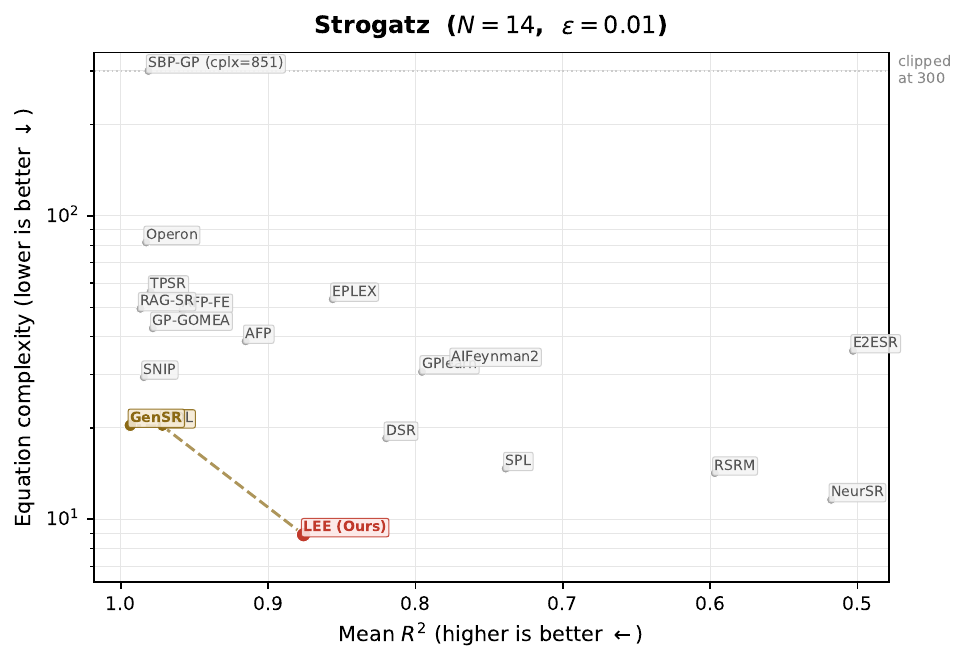}
\caption{Strogatz, $\epsilon{=}0.01$}\end{subfigure}\hfill
\begin{subfigure}[b]{0.48\textwidth}\includegraphics[width=\textwidth]{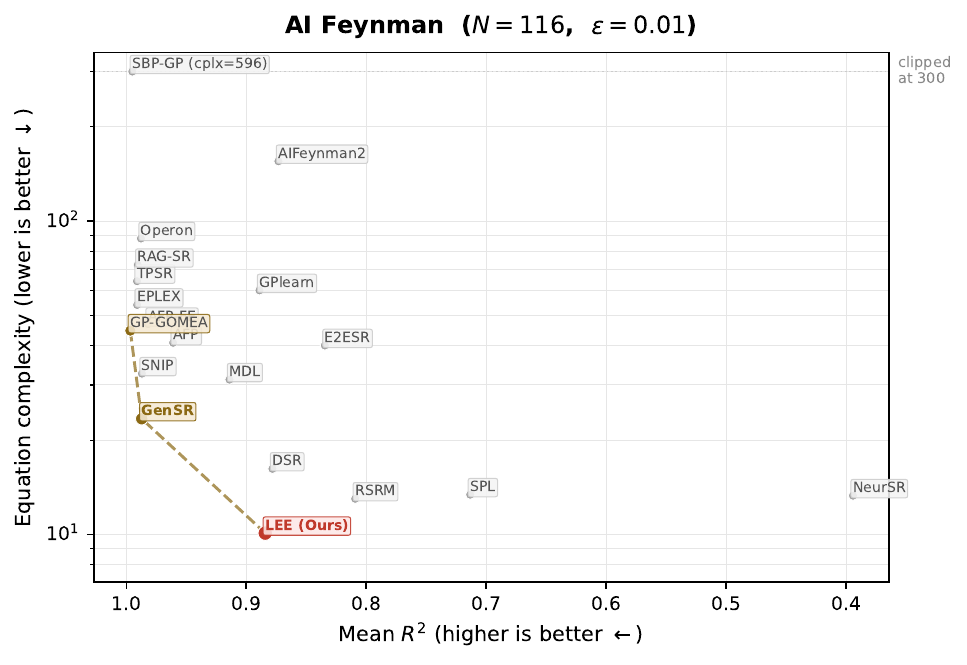}
\caption{Feynman, $\epsilon{=}0.01$}\end{subfigure}
\caption{\textbf{Pareto frontiers on clean and mildly noisy data.}}
\label{fig:pareto-clean}
\end{figure}

\begin{figure}[h]
\centering
\includegraphics[width=0.55\textwidth]{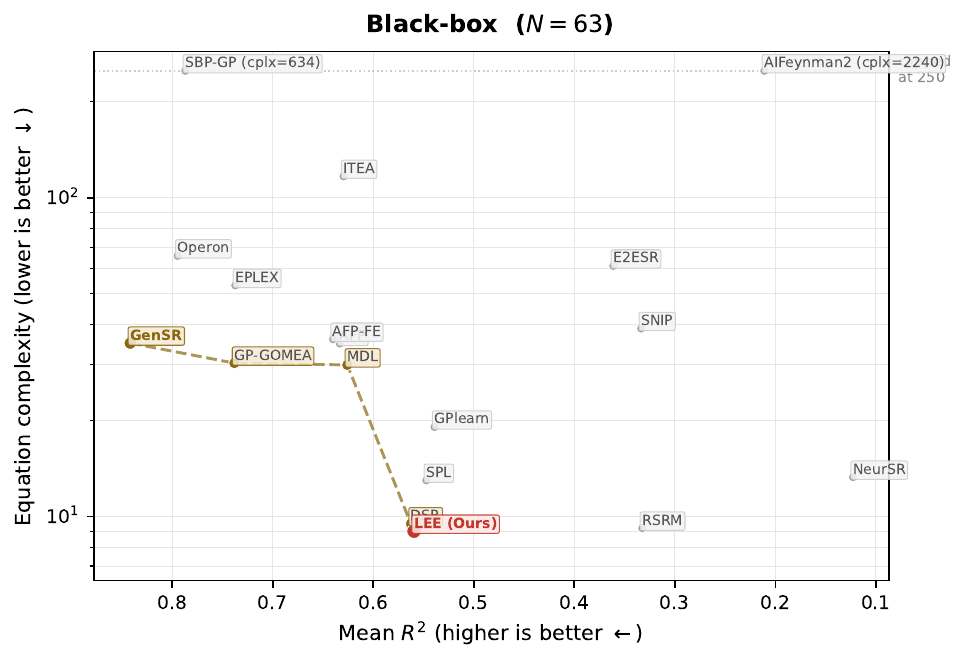}
\caption{\textbf{Pareto frontier on the black-box benchmark} (test $R^2$ vs.\ complexity). LEE anchors the low-complexity end; GenSR anchors the high-accuracy end. Several one-shot neural baselines (SNIP, E2ESR, NeurSR) are dominated in both dimensions.}
\label{fig:pareto-bb}
\end{figure}

The main text (Fig.~\ref{fig:pareto}) shows the Strogatz and Feynman Pareto frontiers at the hardest noise level $\epsilon{=}0.1$ for clarity. Figure~\ref{fig:pareto-clean} reproduces the same diagram at $\epsilon{=}0$ and $\epsilon{=}0.01$; Fig.~\ref{fig:pareto-bb} adds the black-box frontier. Across every setting, LEE sits in the low-complexity corner.

\end{document}